\pdfoutput=1
\documentclass[10pt,leqno]{amsart}
\usepackage{indentfirst}
\usepackage{natbib}
\renewcommand{\cite}{\citep}
\usepackage{xcolor,caption}
\usepackage{hyperref}
\hypersetup{colorlinks=true,linkcolor=black,citecolor=black,filecolor=black,urlcolor=black}
\date{}
\usepackage{latexsym}
\usepackage{amsmath}
\usepackage{booktabs}
\usepackage{tabularx}
\usepackage{amssymb}
\usepackage{longtable}
\usepackage{listings}
\usepackage{placeins}
\usepackage{tikz}
\usetikzlibrary{arrows.meta, positioning, shapes.geometric, fit}
\tikzset{
  box/.style={
    draw,
    rounded corners,
    align=left,
    inner sep=6pt,
    font=\small,
    minimum width=2.7cm,
    minimum height=0.9cm
  },
  decision/.style={
    draw,
    diamond,
    aspect=2,
    align=center,
    inner sep=2pt,
    font=\small
  },
  line/.style={
    -{Latex[length=2mm]},
    thick
  }
}

\usepackage[T1]{fontenc}

\usepackage[utf8]{inputenc}

\usepackage{microtype}

\usepackage{inconsolata}

\usepackage{graphicx,adjustbox}
\BeforeBeginEnvironment{tabular}{\begin{adjustbox}{max width=\linewidth}}
\AfterEndEnvironment{tabular}{\end{adjustbox}}
\AtBeginEnvironment{longtable}{\footnotesize\setlength{\tabcolsep}{3.5pt}}
\usepackage{etoc}

\title{EduBehaviors:\\ Assertion-based Schemas for Auditable Coding of Educational Dialogues}

\author[Bernado]{Julian Bernado\textsuperscript{*}}
\author[Ribeiro]{Ana Trindade Ribeiro}
\author[Beberman]{Xander Beberman}
\author[Loeb]{Susanna Loeb}

\dedicatory{\normalfont SCALE Initiative, Stanford University}

\thanks{\textsuperscript{*}Corresponding author:
  \href{mailto:jbernado@stanford.edu}{\texttt{jbernado@stanford.edu}}}

\begin{document}
\baselineskip=16pt
\begin{abstract}
Large language models have enabled rapid deployment of language classification and analysis, allowing users to describe complex constructs in natural language, such as pedagogical moves or expressions of learning and engagement, and label them in conversational datasets such as classroom or tutoring transcripts. 
However, because LLM reasoning is opaque, we have no verifiable, mechanistic insight into \textit{why} a model chose a label for an utterance, and output differences over time or across models challenge replication. We introduce the \textit{EduBehaviors} framework to generate interpretable, reliable, and scalable construct annotations in educational dialogue data that retain human agency while leveraging the strengths of generative AI, encoders, and machine learning models. We evaluate the framework on the TalkMoves dataset, predicting the Teacher TalkMoves labels. Our best configuration results in a macro-F1 of 0.764 and 0.756 Cohen’s $\kappa$, demonstrating improvements over direct prompting approaches and showing competitiveness with fine-tuned classifiers while gaining the benefit of greater transparency and comparability. To facilitate adoption and operationalization of the \textit{EduBehaviors} framework in educational dialogue data by other researchers, we also release the \textit{EduBehaviors Toolkit} with two complementary open-source tools based on the TalkMoves application.
\end{abstract}
\maketitle

\etocdepthtag.toc{main}

\section{Introduction}


Large language models (LLMs) are increasingly being used for text and discourse annotation, with studies showing that they can match or exceed crowdworker performance on some classification tasks at substantially lower cost \citep{gilardi-etal-2023-chatgpt, ziems-etal-2024-large}. This development creates new opportunities for large-scale, theoretically grounded analysis of conversational data, especially in education research, where teacher-student talk has long been treated as an important source of evidence about instructional quality and student learning \citep{michaels-etal-2008-deliberative,demszky-hill-2023-ncte}.

Using LLMs for annotation still requires researchers to define constructs, develop coding procedures, and validate that model outputs capture the intended distinctions \citep{pangakis-etal-2023-validation,reiss-2023-reliability,halterman-keith-2025-codebook}. These activities require expertise and iterative examination of examples, adding to the work involved in producing usable measures. When an LLM assigns a construct label directly, disagreements with human coders can be difficult to diagnose: they may reflect model error or biases, ambiguous evidence, or insufficiently specified construct definitions. Prompt refinement is especially challenging when reference labels are sparse or unavailable, and existing optimization methods typically require explicit feedback signals or labeled development data \citep{he-etal-2025-prompting,pryzant-etal-2023-automatic}.

Evidence from educational discourse illustrates these annotation challenges. Using the TalkMoves dataset \citep{suresh-etal-2022-talkmoves}, \citet{vanacore-kizilcec-2025-instructional} found that the strongest tested model--prompt combination achieved only moderate overall agreement with expert coders ($\kappa=0.58$), despite using a human-validated coding manual and few-shot examples. Related studies also report difficulties with abstract constructs and judgments requiring nuanced interpretation \citep{liu2025qualitative,thomas2025leveraging}. These findings motivate evaluating both the accuracy of construct labels and the intermediate judgments used to produce them.

We introduce EduBehaviors, a framework that combines LLMs with traditional NLP classifiers and machine Learning methods to grant researchers greater oversight of construct definitions and annotation output that is replicable and can scale vertically (related corpora) and horizontally (related constructs). Our framework leverages LLMs' generative capabilities to develop a collection of \textit{assertions} that represent linguistic markers and behaviors, coded as true or false, which can be observed in the corpus and relate to a given construct of interest, and annotate a subset of the corpus for human inspection. After eliminating or refining low-quality assertions, the researcher can use the human-inspected LLM annotations to fine-tune independent, encoder-based classifiers for each remaining assertion, enabling replicability and reuse on related datasets. The LLM- or encoder-annotated assertions can then be used to infer the construct of interest from either rule-based heuristics or, when construct labels are available for some subset of the data, by training a statistical model. Furthermore, a corpus annotated for a sufficiently large collection of assertions that cover a wide array of linguistic markers and behaviors lowers the cost for researchers to label and explore new constructs and definitions, inspect human- or LLM-labeled construct definitions, or even analyze similarities and differences between two constructs in a labeled dataset.


\subsection*{Contributions} Our primary contribution is the \emph{EduBehaviors} framework, centering human decisions and interpretation in LLM-assisted coding of educational conversations at scale. 

The framework exposes both the behavioral judgments used as evidence and the rule or model that maps those judgments to a construct label. This structure allows researchers to examine the evidence used in a classification, revise individual assertions, and assess how those assertions contribute to the final label. It also supports reuse: an assertion annotated once in a corpus can inform multiple construct labels, and assertion annotations can be used to train smaller classifiers for subsequent annotation. These features are intended to reduce repeated development and annotation work, as well as promote replicability of construct measurements.

We evaluate the framework on the expert-labeled TalkMoves dataset \citep{suresh-etal-2022-talkmoves}, comparing its predictive performance with direct LLM prompting and examining agreement across LLM annotators. To facilitate the framework adoption, we also release two open-source tools: EduBehaviors-Studio, an interface for developing and refining assertion-based schemas, and EduBehaviors-kit, an encoder-based annotation pipeline modeled after the TalkMoves dataset.

\section{Related Work}

\subsection{Developing annotation criteria through theory and data}

Operationalizing educational constructs requires connecting theoretical definitions with evidence that can be identified in conversational data. Recent work has applied LLMs to classroom and tutoring discourse constructs at scale \citep{long-etal-2024-evaluating,moreau-pernet-etal-2024-classifying}, extending earlier supervised approaches \citep{demszky-etal-2021-measuring,suresh-etal-2022-talkmoves} and informing analyses of emerging human-AI tutoring systems \citep{wang-etal-2025-tutor}. 

Research on LLM-assisted codebook development approaches this task through a combination of educational theory, empirical examples, and human refinement \citep{zambrano2026codebook}. Complementary work on concept induction helps researchers discover patterns in unstructured text and articulate explicit criteria for identifying them \citep{lam2024lloom}. Recent systems support mixed-initiative codebook discovery \citep{xiong-etal-2025-co} and orchestrated AI-assisted annotation \citep{hedley-etal-2026-sandpiper}. Together, these approaches position LLMs as resources for developing and revising operationalizations through engagement with both theory and data.

This development process is inherently iterative: applying criteria to examples can expose ambiguities and reshape the criteria themselves. Research on LLM-assisted evaluation describes this interaction as \emph{criteria drift}, emphasizing that defining criteria and examining their applications are interdependent activities \citep{shankar2024validators}. These ideas inform both the curation of corpus-derived assertions and the researcher-guided refinement of schema-derived assertions in EduBehaviors.

\subsection{Decomposition and interpretable construct modeling}

Once annotation criteria have been articulated, a related question is how to organize them into an inspectable prediction process. Decomposed prompting and fine-grained evaluation share the principle that complex tasks or broad judgments can be separated into components that are easier to specify, assess, and refine individually \citep{khot-etal-2023-decomposed,ye-etal-2024-flask}. This principle motivates treating behavioral assertions as distinct annotation targets.

Concept bottleneck models connect these intermediate judgments to final predictions by making human-understandable concepts an explicit part of the predictive process \citep{koh2020concept}. Extensions to text classification further demonstrate how LLMs can help identify and measure such concepts \citep{ludan2023textbottleneck}. Together, decomposition and concept-based prediction provide foundations for separating the identification of behavioral evidence from its interpretation as an educational construct. This separation also connects prediction to the iterative development process described above, making the intermediate criteria available for researcher inspection and revision.

\subsection{Reusable models and educational research infrastructure}

Making annotation components reusable requires ways to translate researcher expertise into supervision at scale. Snorkel \citep{ratner-etal-2020-snorkel} demonstrated the use of labeling functions defined by subject matter experts with generative models, combining noisy outputs into training labels to accelerate discriminative model development. Generative and reasoning capabilities make LLMs a powerful tool for classification of hard-to-define concepts and relationships, with distillation approaches emerging to support applications that require the reliability, low latency, and low cost at scale afforded by smaller classification models~\citep{agrawal2025rationale,borchers2025distillation}. The EduBehaviors framework relates to these approaches by using LLMs to develop and annotate assertions based on the corpus and a construct coding schema or preexisting construct labels to develop encoders. This approach enables the researcher to inspect and individually accept or reject assertions, affording more control and understanding of the linguistic markers and behaviors that can be labeled in the corpus and that relate to the construct of interest, as assertions describe distinct behaviors whose predictions serve as features for construct modeling.

Shared infrastructure makes such models available beyond the studies in which they were developed. Edu-ConvoKit supports this goal by integrating educational conversation analysis tools, including established models for focusing questions, accountable talk moves, and conversational uptake \citep{wang2024educonvokit}. It provides complementary resources to EduBehaviors' emphasis on reusable behavioral assertions. Viewed together, these literatures connect researcher-guided operationalization, interpretable prediction, and efficient deployment---the three concerns linking EduBehaviors-Studio and EduBehaviors-kit.

\section{The EduBehaviors Framework}
Consider a set of utterances $\mathcal{U}$ and the task of annotating each utterance $u\in \mathcal{U}$ with a label $\ell \in \mathcal{L} = \{1, \ ..., \ k\}$. Now, let an \textbf{assertion} $A$ be a binary-valued function $A: \mathcal{U} \to \{0,1\}$ on the set of utterances. Then, a \textbf{decomposable schema} $S = (\mathcal{A}, R)$ is a set of assertions $\mathbf{A} = (A_1, ..., A_n)$ along with a rule $R: \{0,1\}^n \to \mathcal{L}$ mapping every combination of assertion values into the label space. Taken together, an assertion set and rule can be composed to form a map $M: \mathcal{U} \to \mathcal{L}$ into the label space defined as $M(u) = R(A_1(u), ..., A_n(u))$. The map $M$ corresponding to a schema $S$ can then be used to predict a label $\ell \in \mathcal{L}$ for every utterance $u\in \mathcal{U}$. As formulated here, a decomposable schema is an example of a \textit{concept bottleneck model}, with concept vector $\mathbf{A}$ and task function $R$ \cite{koh2020concept}. In this section, we describe the \textit{EduBehaviors} framework: a modular, transparent, cost-sensitive approach to annotating a construct in educational discourse data.

\subsection{Generating assertions}
For our decomposable schema $S$ to be interpretable, we need its assertions to correspond to human-readable, directly observable aspects of utterances. For example, "utterance is spoken by a student," or "utterance includes a number" are viable aspects that can be directly annotated in the corpus using a language model. The behaviors encoded by assertions can be human-prescribed or LLM-generated. The EduBehaviors framework uses two assertion-generation steps: generate assertions that describe the corpus of annotation, then generate assertions that decompose the construct of interest.

\par \textbf{Corpus-derived assertions} are generated directly from the corpus, with no grounding in a specific downstream task. In a pedagogical dialogue setting, this may include assertions describing an utterance's pedagogical function ("utterance poses a new exercise"),  intended recipient ("utterance is directed at a student"), relationship to other utterances ("utterance corrects a mistake"), and any other dichotomous aspect of an utterance. The corpus-derived assertions form the construct-independent set of assertions that can be used to annotate \textit{any future construct of interest} on the corpus. 

\par \textbf{Construct-derived assertions} are generated directly from a description of the construct of interest with no grounding in the corpus text. These assertions should correspond to observable aspects of one or more construct labels. For example, if the task of interest is identifying utterances expressing curiosity in tutoring sessions, a natural assertion might be "utterance asks a question" or "utterance expresses interest in lesson." These assertions may be used as evidence that an utterance expresses curiosity, but break the concept down beyond the more subjective annotation "utterance expresses curiosity."

\par For assertions to be useful, their annotation in the dataset should be minimally ambiguous: any two unprepared human annotators should highly agree on which utterances an assertion should be true-valued. While assertion annotations can be reviewed directly, managing a library of reusable assertions requires effective triage of assertions for rewriting. To get a proxy for assertion faithfulness, we annotate each assertion using a panel of LLM annotators, then calculate inter-rater reliability for each assertion. Although models may share systematic errors even in this scenario~\citep{dell2026measurement}, our approach makes measurement judgment more inspectable and allows us to identify cases in which assertions may need to be written differently or discarded if they fail to pass a minimum agreement bar (e.g. Krippendorff's $\alpha \geq 0.5$).

\subsection{Defining a rule}
Given an assertion set $\mathcal{A} = \{A_1, ..., A_n\}$, the next step is to specify or learn the mapping rule $R: \{0,1\}^n \to \mathcal{L}$ must be decided. However, since our rule only operates on the binary assertion values, this amounts to a classification task with binary covariates. As such, transparent linear classifiers such as logistic regression may be used to maximize interpretability of the resulting schema. Otherwise, more flexible procedures like a random forest or fine-tuned neural network might be used to capture complex relationships between assertions and improve predictive accuracy.

\par Because the rule definition step is a standard tabular supervised learning problem, our framework has two desirable properties not present in direct LLM annotation. The first is \textbf{calibration}: calibrated classification probabilities are not straightforward to derive from LLM annotations \cite{ahtisham2026llmreasoningpredictsmodels}, whereas our framework's classifier may be calibrated with standard techniques, enabling utterance-level triage of uncertain predictions. The second is \textbf{tunability}: precision and recall are competing goals that may each be desired in different situations, and while prompt-based LLM annotation gives no reliable control over this tradeoff, many supervised classifiers have thresholds that directly control it.

The \textit{EduBehaviors} framework described above outlines a procedure for annotating educational conversational data that incorporates the natural language interface of direct LLM prompting into a supervised learning framework. By doing so, we increase the interpretability and reproducibility of our classifications and produce LLM artifacts that can be re-used across many tasks.

\section{EduBehaviors Toolkit}
In this section we describe the EduBehaviors Toolkit, consisting of \textbf{EduBehaviors-Studio}, an LLM-assisted UI for generating construct-derived assertions, and \textbf{EduBehaviors-kit}, a Python package allowing researchers to apply the EduBehaviors framework to their own dataset to generate annotations.

\subsection{EduBehaviors-Studio}
EduBehaviors-Studio\footnote{\url{https://github.com/scale-nssa/edubehaviors-studio/}} is a self-hosted, bring-your-own-key web app allowing researchers to generate a set of construct-derived assertions, annotate them on a dataset of choice, then iterate on these assertions until they adequately reflect a construct of interest.  The app is defined by a single workflow (ref. appendix section~\ref{sec:appendix studio}), where researchers: (1) \textbf{describe} the construct of interest, its possible values, and whether it manifests in tutor utterances, student utterances, or both; (2) \textbf{approve}, reject, or edit a set of LLM-generated, construct-derived assertions based on that description; (3) \textbf{review} utterances to check whether the assertions' implied mapping aligns with their intuition of the construct, and whether the annotated values are faithful; (4) \textbf{revise} assertions or their mapping into a final construct label based on disagreements observed in review; and (5) \textbf{repeat} until the assertions consistently identify the construct in a conversational dataset by the reviewer's validation criteria.
\par Our review process centers our distinction of \textbf{definitional errors} from \textbf{predictive errors}. When encountering a construct classification that a reviewer disagrees with, the reviewer can observe the assertions that fired for a given utterance. If the reviewer is in agreement with the LLM's judgement about the assertions for that utterance yet still disagrees with the final label, this exposes a gap in the measure's definition rather than a flaw in the LLM's annotation. At this point, the reviewer can rephrase an existing assertion, add a new assertion, or revise the aggregation rule. If the rule $R$ is the only change, the counterfactual annotation can be immediately derived with no additional LLM calls. If there is a change to assertions, then the additional costs corresponding to schema refinement are reduced since unchanged assertions need not be re-annotated. These properties give actionable next steps when labels diverge from reviewer judgement beyond just ``make the prompt better.'' 

\subsection{EduBehaviors-kit}

We introduce \texttt{EduBehaviors-kit}, a combination of (I) an open-source, \texttt{pip}-installable Python package accessible on GitHub\footnote{\url{https://github.com/scale-nssa/edubehaviors-kit/}} and PyPI\footnote{\url{https://pypi.org/project/edubehaviors-kit/}}, and (II) a collection of pretrained open-weights encoder models and training data accessible on HuggingFace\footnote{\url{https://huggingface.co/collections/StanfordSCALE/assertions}}, in order to provide an easy-to-use system for researchers to apply the methods described in this paper to their own work (ref. appendix section~\ref{sec:appendix package}). \texttt{EduBehaviors-kit} was created with two main goals in mind:

\par \textbf{Democratization of educational text classification methods.}
\texttt{EduBehaviors-kit} is intended to make lightweight behavior annotation and model training as simple as possible for researchers with little to no experience with language model-based text annotation or classification.
The package allows novice users to easily replicate a pipeline similar to the one described in this paper;
annotating a dataset with all available assertion models and counts of default words, splitting into train/test sets, training and evaluating an interpretable classification model is as simple as calling \texttt{ClassificationPipeline(data, words="all", assertions="all")}.

\par \textbf{Consistent, efficient, and open models.}
By using pretrained encoders to annotate data instead of proprietary language models, we can improve both the reliability and cost of annotation. As such, we provide pretrained classifiers for 102 of the bottom-up and top-down assertions used in this paper on our HuggingFace repository. SetFit \cite{setfit} models are trained on LLM annotations of teacher utterances from a subset of the TalkMoves dataset; training data, performance metrics, and model hyperparameters are publicly available. These lightweight models can be run on a laptop, or cheaply on computing systems. Unlike proprietary language models, encoders also allow for fully reproducible annotation results. By sharing model weights, data and training information, we encourage other users of this work to create their own assertion classifiers or improve upon ours. Details on model training and model evaluation metrics are shown in appendix section \ref{sec:appendix package models}.

\subsubsection{Package structure}
The package contains four main objects:
\begin{enumerate}
    \item \texttt{WordAnnotator}: Annotates data with counts or appearances of words in a list.
    \item \texttt{AssertionAnnotator}: Annotates data using the available assertion encoder models
    \item \texttt{standard\_classifier}: Instantiates a logistic regression classifier with reasonable hyperparameters similar to that used in the analysis for this paper.
    \item \texttt{ClassificationPipeline}: Creates and runs an end-to-end pipeline; annotates data, creates train/test sets, and trains a classification model.
\end{enumerate}
See appendix section \ref{sec:appendix package usage} for usage examples of these objects. Full documentation of these objects, including lists of available assertion classifiers and default words, and examples of how to create an end-to-end classification pipeline, are available on GitHub.

\section{Performance}


\subsection{Task}
\label{sec:task}

The TalkMoves Dataset is a set of 566 human-generated transcripts of mathematics lessons taking place in real classrooms \cite{suresh-etal-2022-talkmoves}. These transcripts were then annotated with the TalkMoves coding manual which describes 10 "talk moves," discursive techniques of pedagogical interest used by either the teacher or student. Appendix~\ref{sec:appendix data} describes the seven Teacher TalkMoves, their distribution, and details on which sample we used (Tables~\ref{tab:data-subsets} and~\ref{tab:data-labels}).

\par In our experiment we focus solely on the seven labels described in the Teacher TalkMoves coding manual. To get an estimate of our classifier performance without annotating the whole dataset we randomly sample 50 sessions, consisting of 14,341 total teacher utterances to annotate. The metrics of interest are Macro-F1 on the gold labels over the 7 classes as well as inter-rater reliability with the gold annotations as measured by Cohen's $\kappa$. We will estimate these metrics using 5-fold session-grouped cross validation and averaging performance on the unseen test set. 

\par The TalkMoves dataset is commonly used as a benchmark for prediction in education settings, allowing us to understand our results in light of existing benchmarks on the dataset. Fine-tuned RoBERTa-base performance as presented in the original TalkMoves paper reached 0.76 macro-F1 with no agreement metrics reported. LLM-prompted performance lies distinctly below the encoder performance; \citet{vanacore-kizilcec-2025-instructional} evaluated the performance of LLMs from three Frontier labs and found a maximum macro-F1 of 0.61 and maximum Cohen's $\kappa$ of 0.58, associated with directly prompting Claude Opus 4.5 with the TalkMoves coding manual and few-shot examples.

\subsection{Experimental Setup}
To benchmark the performance of the EduBehaviors framework on the Teacher TalkMoves task, we  generate both corpus-derived and construct-derived assertions once, annotate them with a variety of language models, then learn an L1-regularized, logistic regression map from the assertion-values into the label space.  We describe the details of this process below.

\par Assertion generation is once-off and shared across all models. For corpus-derived assertion generation, we prompted Claude Fable 5 with TalkMoves transcripts and the task of describing in general terms \textit{what is happening} in each utterance.  Across many transcripts, this resulted in thousands of assertions. We took two approaches to de-duplication. First, we fed the utterance-level annotations into the same model to  generate 61 assertions summarizing the range of behavior described in the utterance-level annotations. In addition, we conducted simple lemmatization (plural and article removal) across the utterance-level annotations, then took the top 50 resulting assertions. After a human-led de-duplication of these two lists, our final list contains 72 corpus-derived assertions. For construct-derived assertions, we prompted Gemini 3.1 Pro with the Teacher TalkMoves section of the TalkMoves coding manual and directly generated 31 assertions with no inclusion of transcripts. Finally, as an automatic source of corpus-derived assertions, we ranked the top 100 words present in a disjoint set of transcripts, then included 100 additional corpus-derived assertions of the form ``The utterance contains the word X." In total, this resulted in 203 total assertions to be used for classification. Prompts are included in Appendix~\ref{sec:appendix prompts}. 

\par Assertion annotation was conducted with a panel of five language models: \textbf{Claude Opus 5}, \textbf{Gemini 3.8 Flash}, \textbf{Gemini 3.5 Flash-Lite}, \textbf{GPT 5.6 Luna}, and \textbf{GPT OSS 120B}. These represent a range of providers, sizes, and includes an open-weights model (GPT OSS). After annotation, cross-model agreement was calculated using Krippendorff's $\alpha$ for each assertion. Model provider and inference settings are described in~\ref{sec:appendix costs}

\par Classification was done by fitting an L1-penalized logistic regression to the final labels based on the gold Teacher TalkMoves annotations. In particular, we fit 7 one-way classifiers corresponding to each possible Teacher TalkMoves label, including ``None." Based on an independent inner-validation loop, each classifier selects a regularization penalty $\lambda$ and selection threshold $\tau$ maximizing validation F1. Then, final labels are selected by taking the \text{arg max} over the resulting probabilities divided by its selection threshold $\tau$, to account for class imbalance. We construct five annotators, leaving out each test session to check for accuracy, then pool performance metrics across the 5 folds.

\par As a consistent baseline, we predict the final labels using only the ``Utterance contains word X" assertions, which are calculated automatically and annotator-independent. Then, we run each annotator in two conditions corresponding to the set of assertions included in the classification step: classifiers learned using only an annotator's corpus-derived assertion values versus classifiers learned using both corpus-derived and construct-derived assertions.

\par After collecting the results for annotating using LLMs, we test the scalability of our method by fine-tuning \texttt{sentence-transformers/paraphrase-mpnet-base-v2} with SetFit on the annotations coming from the two annotators with highest agreement. We then deploy a classifier trained on the original 50 sessions to 495 entirely unseen sessions and report classification metrics.

\subsection{Results}

In this section we describe the results of our assertion annotation, then the resulting classifications.

\subsubsection{Annotation Results}

\begin{figure}[t]
    \centering
    \includegraphics[width=0.8\columnwidth]{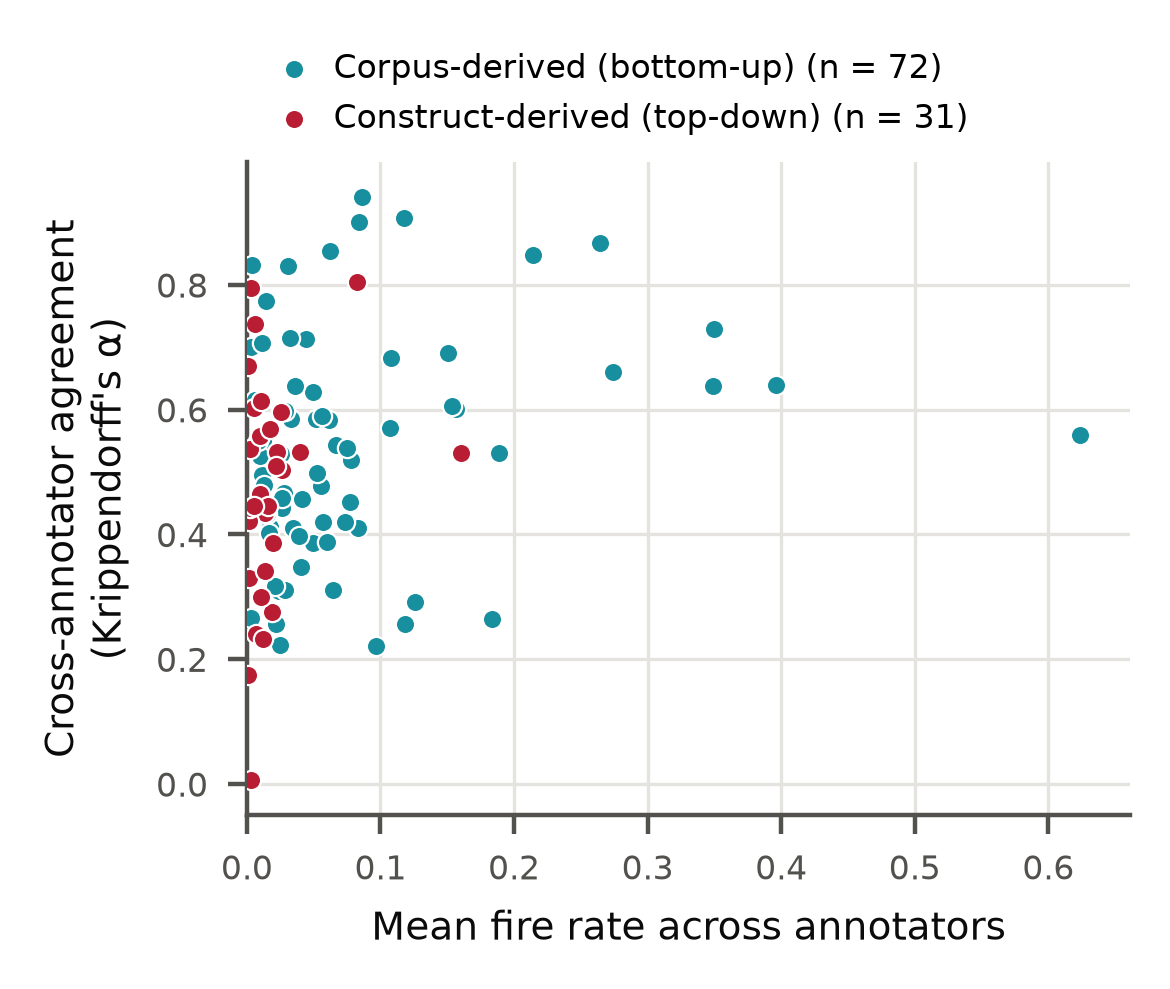}
    \caption{{Fire rate (fraction of teacher utterances where that assertion is true) versus cross-model agreement (Krippendorff's $\alpha$ over the five annotators) for each assertion. Fire rate is averaged across the five annotators. Corpus-derived (bottom-up) assertions are teal; construct-derived (top-down) assertions are red.}}
    \label{fig:fire-agreement}
\end{figure}

The mean prevalence and cross-model agreement of each assertion can be seen in Figure~\ref{fig:fire-agreement}. Overall, cross-model agreement is moderate (median = 0.52) and higher among the corpus-derived assertions (median = 0.53) than the construct-derived assertions (median = 0.47). Fire rate also varies between the two categories, with the corpus-derived assertions appearing more than those that were construct-derived. The two most-agreeing annotators were Claude Opus 5 and GPT 5.6 Luna. Agreement by assertion source and between each pair of annotators is reported in Appendix~\ref{sec:appendix agreement} along with the effect of restricting donwstream predictions to only use assertions with cross-model agreement greater than some threshold (Figure~\ref{fig:agreement-sweep}). Annotation costs (ranging from \$6.74 to \$193.49) are fully reported in Appendix~\ref{sec:appendix costs}, and every assertion with its agreement and fire rate is listed in Appendix~\ref{sec:appendix assertions}.

\subsubsection{Classification Results}

The performance of each annotator with access to all assertions (words, corpus-derived, and construct-derived) is in Table~\ref{tab:best-per-annotator}. Overall, we find that adding corpus-derived, then construct-derived assertions contribute to predictive accuracy as measured by Macro-F1 and Cohen's $\kappa$. Claude Opus 5 attains the highest Macro-F1 of 0.764 as well as the highest Cohen's $\kappa$ of 0.756. Individual $F_1$ scores are shown per label, demonstrating significant variability. All of our annotators outperform the direct prompting baseline (Macro-F1 of $0.61$ and Cohen's $\kappa$ of 0.58). Furthermore, our encoders which annotate with no per-call API charges are competitive with the direct prompting baseline on Macro-F1 and superior when measured by Cohen's $\kappa$. The gap between our encoders, trained with no context prior to the utterance under annotation, and our LLM annotators that annotate a whole session at a time is concentrated in the Restating and Revoicing labels. All details on every configuration and detailed classifier metrics are in Appendix~\ref{sec:appendix classification}, and for the encoders in Appendix~\ref{sec:appendix package}. We also include alternative applications of our framework, such as interpretation of predictive features for related constructs and error analysis, and additional prediction calibration and data efficiency analyses in appendices ~\ref{sec:appendix classification}, ~\ref{sec:appendix interpretability}, and ~\ref{sec:appendix errors}.


\begin{table*}[t]
\centering
\small
\setlength{\tabcolsep}{4pt}
\resizebox{\textwidth}{!}{%
\begin{tabular}{llrrrrrrrrr}
\toprule
& & \multicolumn{2}{c}{Overall} & \multicolumn{7}{c}{Per-class $F_1$} \\
\cmidrule(lr){3-4} \cmidrule(lr){5-11}
Annotator & Test set & Macro-$F_1$ & $\kappa$ & None & KET & GSR & PFA & PFR & RES & REV \\
\midrule
Claude Opus 5 & 50 sessions & \textbf{0.764} {\footnotesize[0.738, 0.786]} & \textbf{0.756} {\footnotesize[0.731, 0.780]} & \textbf{0.941} & \textbf{0.683} & \textbf{0.662} & \textbf{0.837} & 0.748 & \textbf{0.843} & 0.631 \\
Gemini 3.8 Flash & 50 sessions & 0.750 {\footnotesize[0.724, 0.771]} & 0.731 {\footnotesize[0.706, 0.756]} & 0.938 & 0.644 & 0.628 & 0.799 & \textbf{0.780} & 0.820 & \textbf{0.638} \\
GPT-5.6 Luna & 50 sessions & 0.728 {\footnotesize[0.701, 0.750]} & 0.722 {\footnotesize[0.697, 0.746]} & 0.932 & 0.642 & 0.561 & 0.822 & 0.769 & 0.790 & 0.581 \\
Gemini 3.5 Flash-Lite & 50 sessions & 0.706 {\footnotesize[0.676, 0.734]} & 0.696 {\footnotesize[0.668, 0.724]} & 0.926 & 0.629 & 0.536 & 0.783 & 0.766 & 0.792 & 0.513 \\
GPT-OSS 120B & 50 sessions & 0.649 {\footnotesize[0.620, 0.673]} & 0.648 {\footnotesize[0.619, 0.673]} & 0.913 & 0.569 & 0.415 & 0.774 & 0.759 & 0.657 & 0.456 \\
\midrule
MPNet encoders (Opus + Luna tuned) & 495 sessions & 0.590 {\footnotesize[0.583, 0.598]} & 0.636 {\footnotesize[0.629, 0.643]} & 0.898 & 0.623 & 0.508 & 0.780 & 0.734 & 0.372 & 0.216 \\
None (words only) & 495 sessions & 0.388 {\footnotesize[0.383, 0.394]} & 0.366 {\footnotesize[0.358, 0.374]} & 0.821 & 0.274 & 0.130 & 0.618 & 0.661 & 0.162 & 0.052 \\
\bottomrule
\end{tabular}%
}
\caption{Classification performance per annotator. $F_1$ and $\kappa$ are pooled out-of-fold over five session-grouped folds on 50 sessions (14{,}341 teacher utterances); brackets are 95\% session-bootstrap intervals; bold marks the best annotator. Every annotator's row uses words, corpus-derived and construct-derived assertions; Table~\ref{tab:fit-grid} has the other configurations. The rows below the rule are trained on those 50 sessions and tested on 495 others (153{,}985 teacher utterances).}
\label{tab:best-per-annotator}
\end{table*}

\begin{figure}[t]
    \centering
    \includegraphics[width=0.8\columnwidth]{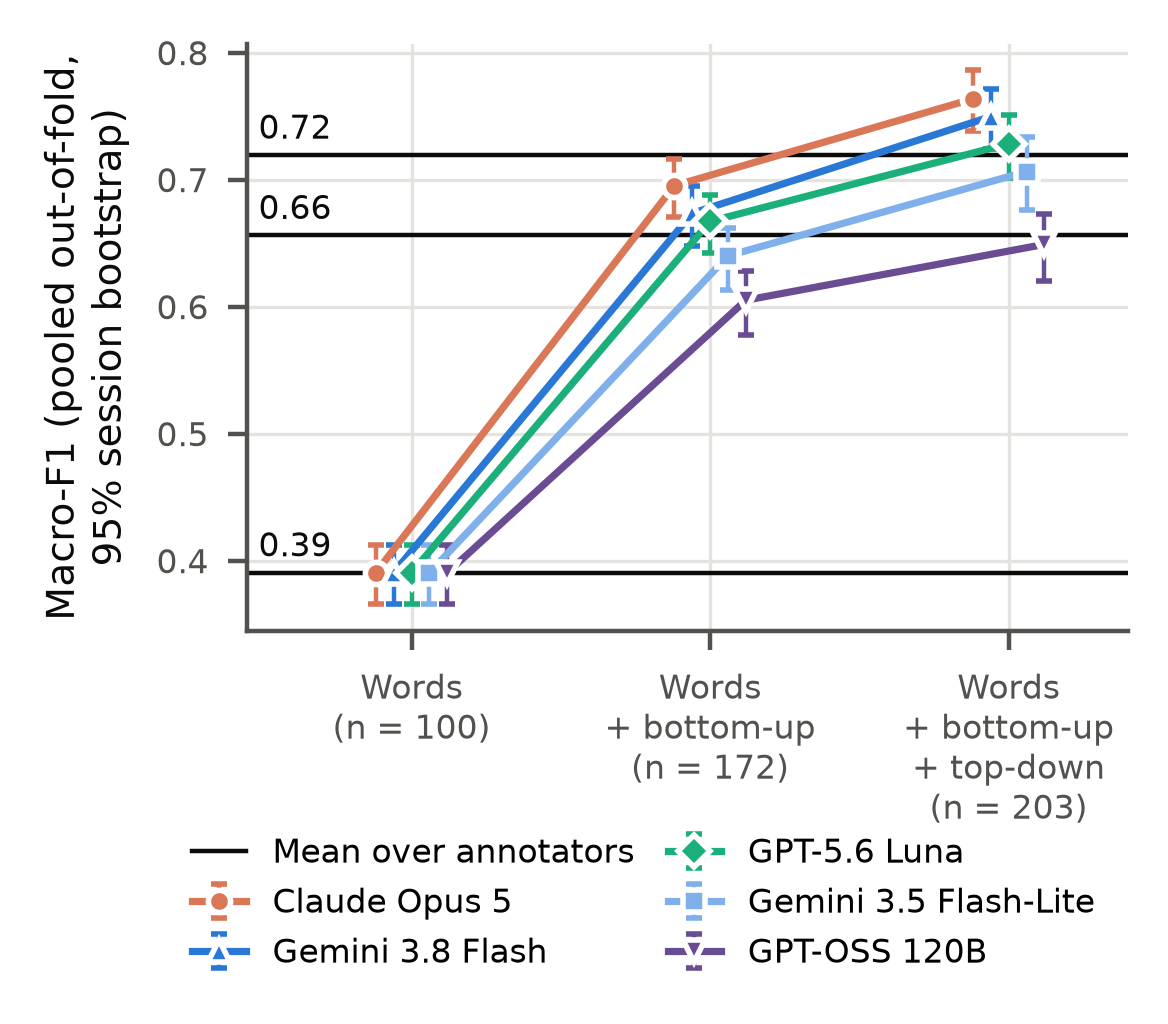}
    \caption{Macro-F1 of each annotator's classifier as assertions are added: words only, then corpus-derived (bottom-up) assertions, then construct-derived (top-down) assertions, without the agreement filter; $n$ is the number of features. Macro-F1 is pooled out-of-fold over five session-grouped folds, with 95\% session-bootstrap intervals; black lines mark each rung's mean over the five annotators. Words only needs no annotation, so it is the same for every annotator.}
    \label{fig:fit-ladder}
\end{figure}

\section{Discussion}

We developed EduBehaviors as an interpretable, reliable, and scalable alternative to traditional codebook development and construct annotation, which is often costly and time-consuming. While developing codebooks and annotating constructs for multiple related teaching practices, we found overlap in linguistic markers and conversational behaviors across constructs, revealing the possibility of modeling hard-to-define constructs based on corpus and construct features and reducing repetitive data annotation that implicitly codes a finite set of observable behaviors. EduBehaviors organizes coding into two steps: identifying specific behaviors and combining those observations into construct labels. For example, identifying whether an educator asks a question can contribute to measures of focusing questions, funneling questions, responses to student mistakes, or relationship building. Other evidence helps distinguish these uses, including what the question asks, whether it follows a student error, and how it relates to the student's preceding contribution. Researchers can reuse the question annotation across these measures while evaluating whether each combination of evidence captures the intended construct.


In the TalkMoves evaluation, the approach generally improved performance over direct LLM prompting, although gains varied across models and metrics. It did not match the performance of the specialized classifier trained on expert-labeled TalkMoves data. These findings establish the approach's predictive performance in one setting. 

Our results also highlight the benefits of the EduBehaviors approach irrespective of classifier accuracy. By separating assertion annotation from rule fitting, we can understand whether a given error was driven by a mislabeled assertion or by insufficiency of the set of assertions and rule. For example, of Claude Opus 5's 1,654 errors, 331 (20$\%$) would be resolved by swapping Opus' own assertion annotations with the majority assertion label for those utterances, 408 errors (25$\%$) would be corrected by swapping with a different annotator on those utterances, and 915 (55$\%$) errors cannot be fixed by any annotator swaps. This allows us to focus our classifier iteration on improving annotation when the former errors occur and adding new assertions or finding a new rule function when the latter error occurs. 

Annotated assertions can also provide training data for smaller classifiers, potentially reducing the time and computational cost of annotating larger samples where appropriate. EduBehaviors-Studio and EduBehaviors-kit support the development and application of these annotation components. Sharing assertion definitions, annotated examples, trained classifiers, and combination rules would give researchers resources they can examine and adapt, reducing the need to develop every measure from the beginning. We hope these shared resources will make the analysis of educational conversations more accessible and help researchers build on one another’s work.


\section{Conclusion}
We introduced EduBehaviors, a framework for annotating educational conversations by identifying specific behaviors and combining those observations into construct labels. This structure allows researchers to inspect and revise individual coding decisions and reuse relevant behavioral annotations across constructs. In the TalkMoves evaluation, the approach generally improved performance over direct LLM prompting, although results varied across models and metrics and remained below those of the specialized classifier trained on expert-labeled data. EduBehaviors-Studio and EduBehaviors-kit provide tools for developing and applying assertion-based schemas. Together, the framework and tools help researchers examine how their measures are constructed and build on existing annotation work, while retaining the need to validate each measure for its intended use.

\section*{Limitations}
\subsection*{Validation of assertion annotations} We acknowledge that validation of LLM-annotated assertions against a human-generated, golden dataset is preferable whenever possible, but recognize the challenges of creating validation sets for dozens of assertions. In the absence of reference assertion labels, cross-model agreement provides evidence of consistency across annotation systems, but cannot exclude shared errors or independently establish construct validity. We caution users of our framework relying on cross-LLM agreement as a proxy for assertion validity to audit assertion annotations even when cross-LLM agreement is high.

\subsection*{EduBehaviors-kit classifiers out of context} At the time of writing, all encoder-based classifiers available through the EduBehaviors-kit are based on TalkMoves corpus-derived assertions, using LLM-annotated labels on the TalkMoves corpus as training and validation. Thus, these models may not be valid for applications that differ in significant dimensions, and their use should be carefully evaluated in such cases.

\subsection*{Framework flexibility and user discretion} Our framework is highly flexible and depends on the user to make sensible choices when implementing it on a new context. Our toolkit application is modeled after the framework demonstration presented in this paper as an accessible implementation, but is not intended to be prescriptive.

\subsection*{Performance against fine-tuned encoders} While we are able to demonstrate that the EduBehaviors framework outperforms directly prompting LLMs in all tested cases that used LLM-annotated assertions, only two models, Claude Opus 5.0 and Gemini 3.8 Flash, produced macro-F1 scores comparable to the original fine-tuned encoder-based classifier trained on the golden dataset annotated by experts. Thus, the investment necessary for creating such datasets may still be a worthwhile one in conditions that require a high degree of confidence in model performance.

\section*{Acknowledgments}
We thank Aaliyah Agyen for excellent research assistance; attendees of the ``National Tutoring Observatory’s Interactive Workshop on AI Annotation and Teacher Analytics'' at the Learning Analytics and Knowledge 2026 and the ``Advancing the Science of Human and AI Tutoring through Shared Infrastructure: A Collaborative Workshop'' at the Festival of Learning 2026, Rene Kizilcec, Kirk Vanacore, John Whitmer, Adam Goldfarb, and the SCALE Initiative team for feedback and comments. We would like to thank Stanford University and Stanford Research Computing for providing computational resources and support that contributed to these research results. This work was supported by the Gates Foundation under grant INV-071948.

\clearpage
\appendix
\etocdepthtag.toc{appendix}
\etocsettagdepth{main}{none}
\etocsettagdepth{appendix}{section}
\etocsettocstyle{\noindent\textbf{Appendix contents}\par\smallskip}{\bigskip}
\etocsetstyle{section}{}{}{\noindent\makebox[1.5em][l]{\etocnumber}\etocname\dotfill\etocpage\par}{}
\tableofcontents

\section{EduBehaviors-Studio Interface}
\label{sec:appendix studio}
\begin{figure*}[t]
    \centering
    \includegraphics[width=0.9\textwidth]{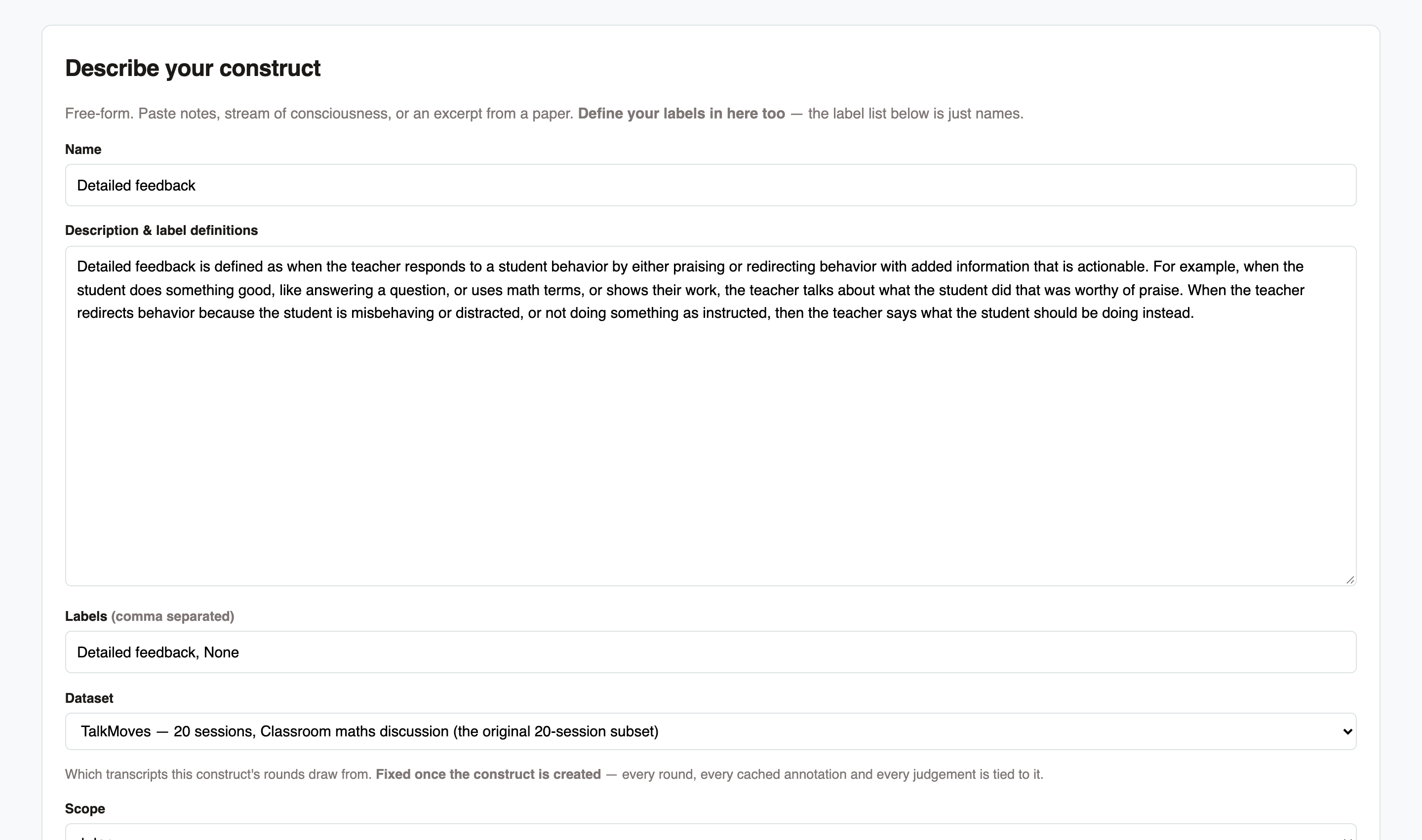}
    \caption{Describe: the user names a construct, defines it and its labels in free text, before this description is turned into a set of assertions.}
    \label{fig:studio-describe}
\end{figure*}

\begin{figure*}[t]
    \centering
    \includegraphics[width=0.9\textwidth]{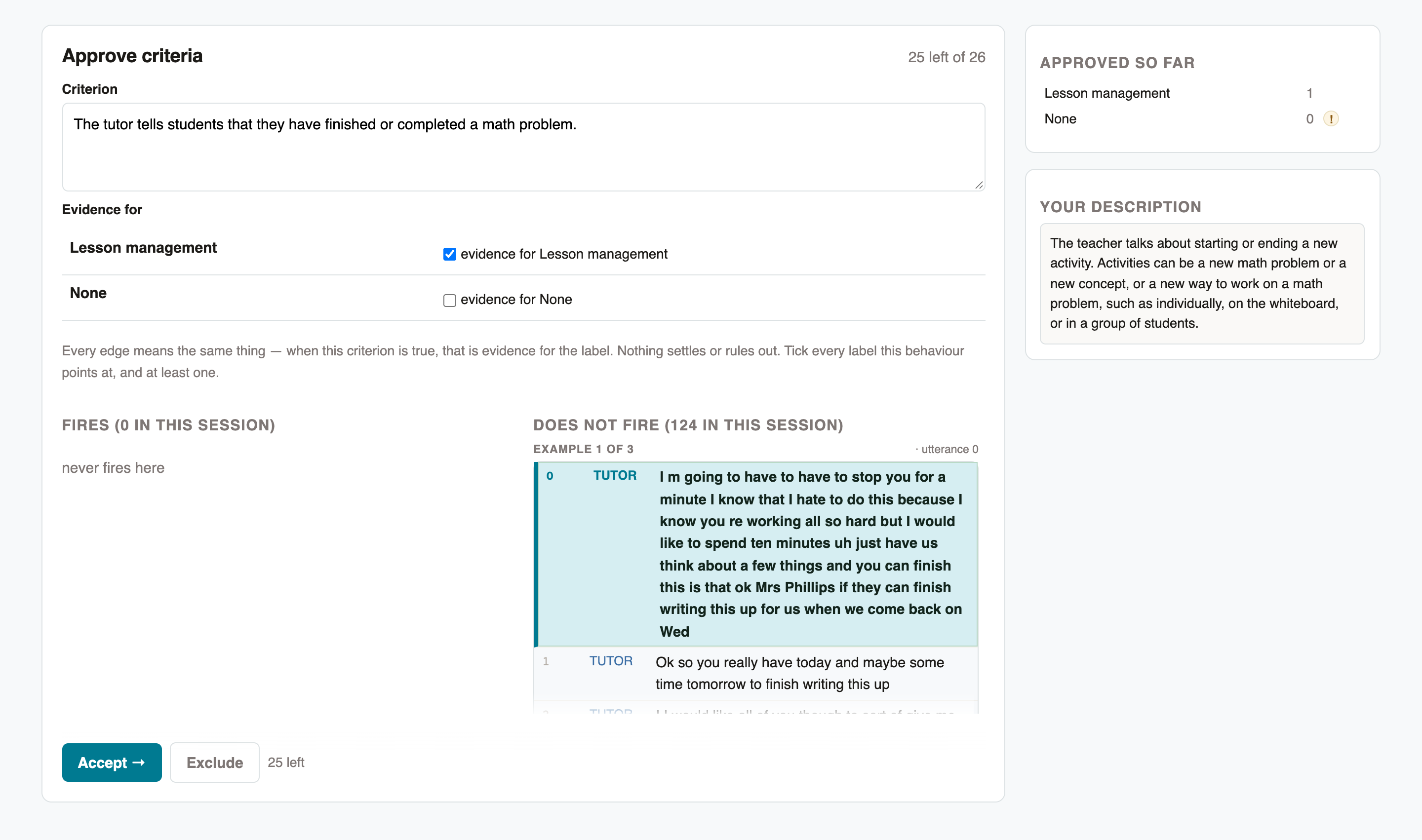}
    \caption{Approve: the LLM-generated assertions are approved or denied, and the label they carry evidence for is determined.}
    \label{fig:studio-approve}
\end{figure*}

\begin{figure*}[t]
    \centering
    \includegraphics[width=0.9\textwidth]{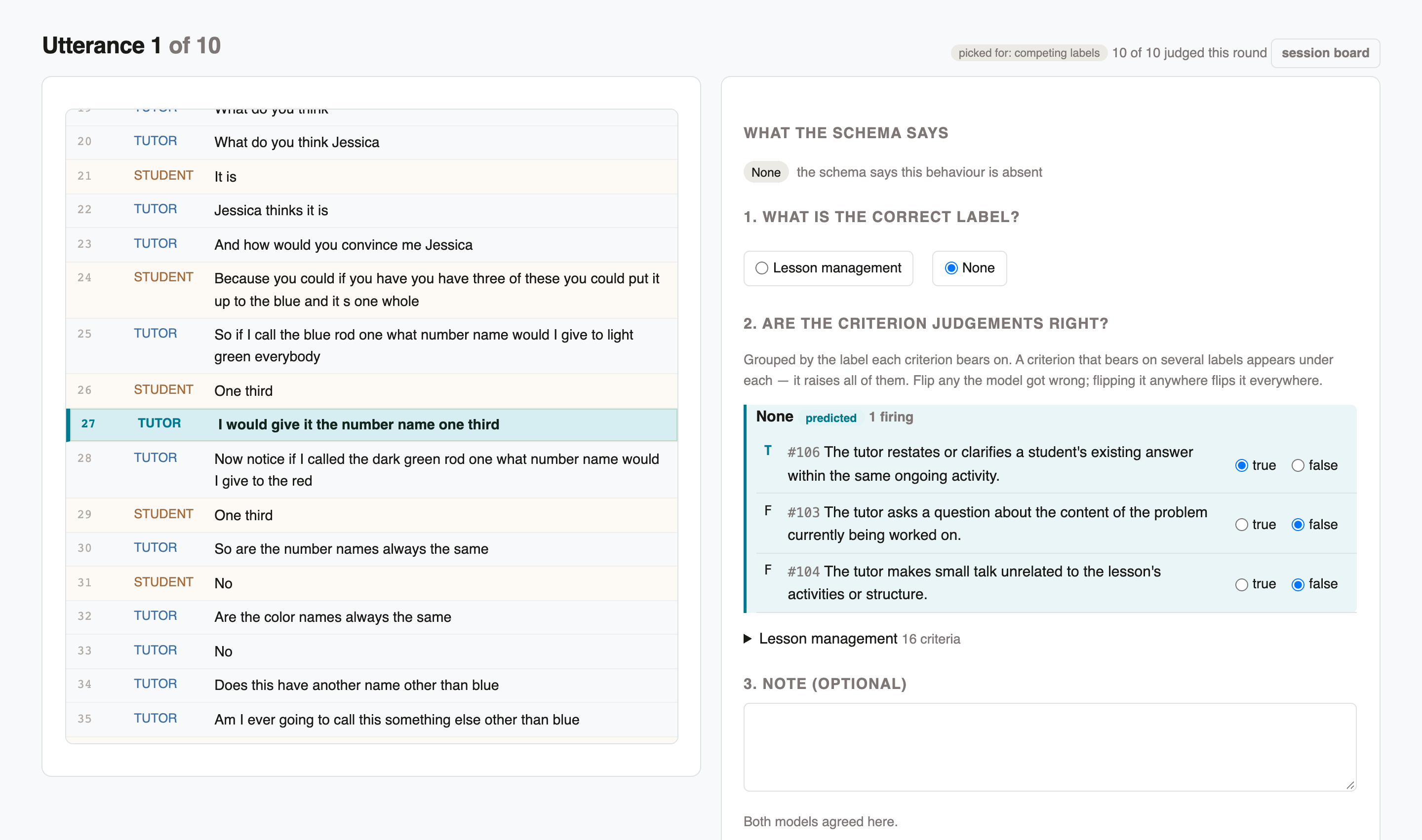}
    \caption{Review: the user goes through utterances in a dataset, seeing whether the schema implied by their assertions holds up with their intuition about the construct.}
    \label{fig:studio-review}
\end{figure*}

Figures~\ref{fig:studio-describe} through \ref{fig:studio-review} show the describe, approve, review loop in the EduBehaviors-Studio app. The aim of EduBehaviors Studio is to provide researchers with a non-technical interface from which they can generate assertions, annotate them, and see whether their assertions speak to the construct they wish to annotate. The web app features LLM suggestions for schema edits, version history tracking, and the ability to export assertion annotations for use with EduBehaviors-kit.

\begin{figure*}[t]
    \centering
    \includegraphics[width=0.9\textwidth]{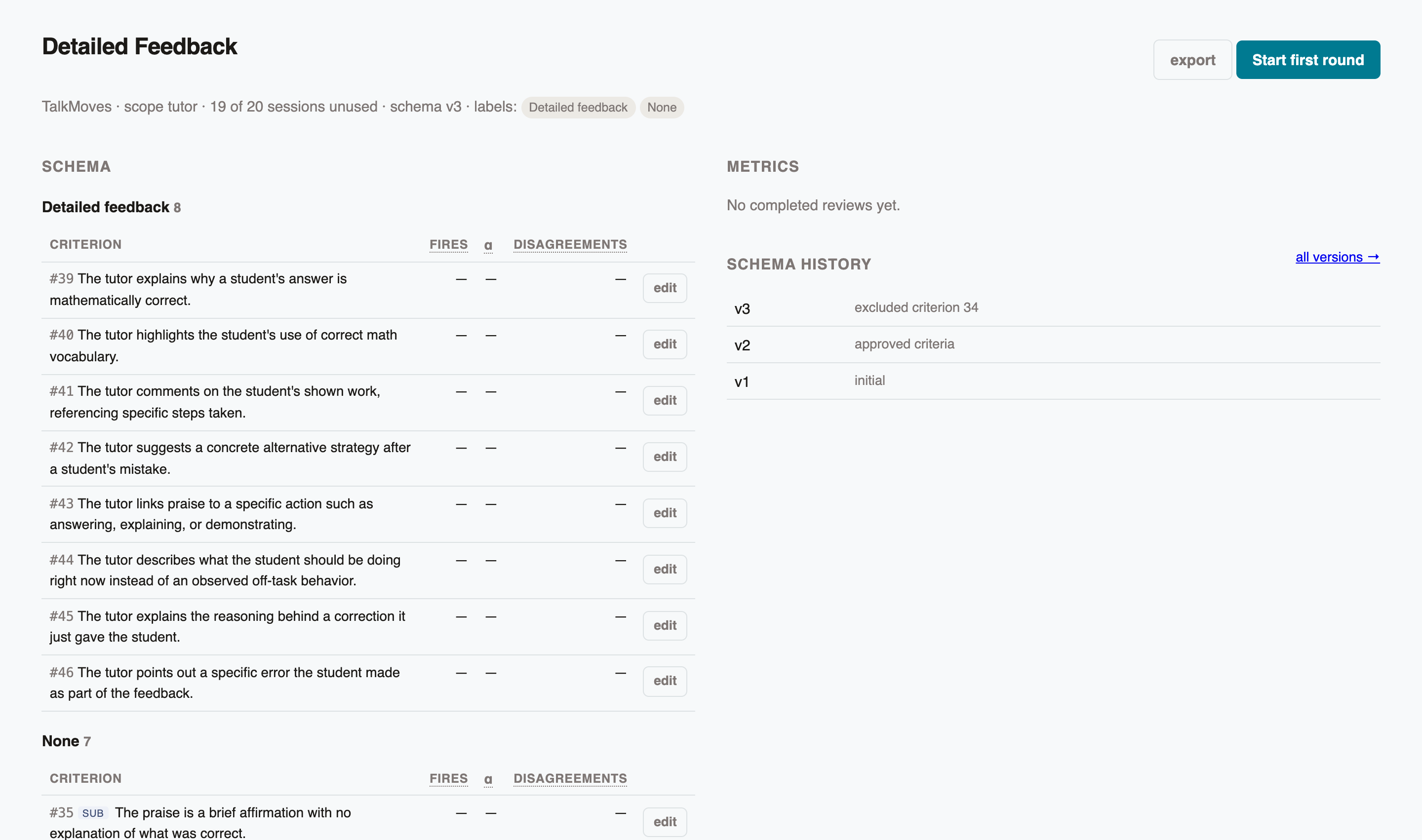}
    \caption{A between-round's review page allowing the user to understand changes they've made to the construct definition and the current agreement they have with its predictions.}
    \label{fig:studio-schema}
\end{figure*}

\FloatBarrier
\clearpage

\section{EduBehaviors-Kit Details}
\label{sec:appendix package}

\subsection{Usage}
\label{sec:appendix package usage}
\definecolor{consolebg}{rgb}{0.97,0.97,0.97} 

The following sections provide introductory usage examples for the three main classes in \texttt{EduBehaviors-kit}. The examples here use the subset of the Talk Moves data included with the package in GitHub. For full usage and reference, refer to the documentation.
\begin{lstlisting}[language=Python,basicstyle=\ttfamily\small,backgroundcolor=\color{consolebg}]
>>> data.sentence.head()
>>> data.sentence.head()
0    My names Ms Villarin I dont know if Ms Ferrant told you about me but I planned with Ms Ferra...
1    So today her and I will be teaching a lesson together again my names Ms Villarin and thatll ...
2    So you can treat me the way you treat her if you ask questions and you have something that y...
3                                                                                               Okay
4                                               So today were going to be doing a really cool lesson
Name: sentence, dtype: str
\end{lstlisting}

\subsubsection{\texttt{WordAnnotator}}
\texttt{WordAnnotator} counts or checks containment of words in a dataset. Use by passing a list of words:
\begin{lstlisting}[language=Python,basicstyle=\ttfamily\small,backgroundcolor=\color{consolebg}]
>>> wordannotator = WordAnnotator(
...     words=["math", "class", "board", "lesson", "question", "okay"],
...     method="count",
... )
>>> wordannotator.annotate(data.sentence)
      word_count__math  word_count__class  ...  word_count__question  word_count__okay
0                    1                  1  ...                     0                 0
1                    0                  0  ...                     0                 0
2                    0                  0  ...                     0                 0
3                    0                  0  ...                     0                 1
4                    0                  0  ...                     0                 0
...                ...                ...  ...                   ...               ...
[1747 rows x 6 columns]
\end{lstlisting}
Or by leaving out the words and method entire during instantiation to get word counts using a predetermined list of 100 words: 
\begin{lstlisting}[language=Python,basicstyle=\ttfamily\small,backgroundcolor=\color{consolebg}]
>>> wordannotator = WordAnnotator()
>>> wordannotator.annotate(data.sentence)
      word_count__you  word_count__to  word_count__the  ...  word_count__give  word_count__he
0                   1               0                0  ...                 0               0
1                   0               0                1  ...                 0               0
2                   6               0                1  ...                 0               0
3                   0               0                0  ...                 0               0
4                   0               1                0  ...                 0               0
...               ...             ...              ...  ...               ...             ...
[1747 rows x 100 columns]
\end{lstlisting}

\subsubsection{\texttt{AssertionAnnotator}}
\texttt{AssertionAnnotator} uses pretrained encoder moders to annotate data. These are lightweight annotators that can run on a laptop. Annotate by choosing a set of assertions, which are typed and described in the documentation:
\begin{lstlisting}[language=Python,basicstyle=\ttfamily\small,backgroundcolor=\color{consolebg}]
>>> assertionannotator = AssertionAnnotator(
...     [
...         "addresses_the_whole_class",
...         "contains_a_time_marker",
...         "is_very_short",
...         "references_classroom_materials",
...     ]
... )
>>> assertionannotator.annotate(data.sentence)
   assertion__addresses_the_whole_class  assertion__contains_a_time_marker  \
0                          False                                  True       
1                           True                                  True       
2                           True                                 False       
3                          False                                 False       
4                           True                                  True       
...                          ...                                   ...

   assertion__is_very_short  assertion__references_classroom_materials  
0                     False                          False              
1                     False                           True              
2                     False                          False              
3                      True                          False              
4                     False                          False
...                     ...                            ...
[1747 rows x 4 columns]
\end{lstlisting}
Or use \texttt{predict\_proba} to get estimated probabilities
\begin{lstlisting}[language=Python,basicstyle=\ttfamily\small,backgroundcolor=\color{consolebg}]
>>> assertionannotator.predict_proba(data.sentence)
   assertion__addresses_the_whole_class  assertion__contains_a_time_marker  \
0                       0.294221                              0.974974       
1                       0.812815                              0.965356       
2                       0.894344                              0.000239       
3                       0.482470                              0.000285       
4                       0.967281                              0.967243     
...                          ...                                   ...

   assertion__is_very_short  assertion__references_classroom_materials  
0                  0.005996                       0.000532              
1                  0.005567                       0.986237              
2                  0.005172                       0.000424              
3                  0.986155                       0.000407              
4                  0.005338                       0.000322  
...                     ...                            ...
[1747 rows x 4 columns]
\end{lstlisting}

\subsubsection{\texttt{ClassificationPipeline}}
\texttt{ClassificationPipeline} conveniently annotates, fits a cross-validated and regularized logistic regression classifier, and predicts on your dataset all in one call:
\begin{lstlisting}[language=Python,basicstyle=\ttfamily\small,backgroundcolor=\color{consolebg}]
>>> pipeline = ClassificationPipeline(
...     data,
...     words="all",
...     assertions=[
...         "addresses_the_whole_class",
...         "contains_a_time_marker",
...         "is_very_short",
...         "references_classroom_materials",
...     ],
...     label_column="label_keeping_everyone_together",
...     text_column="sentence",
...     word_method="contains",
...     assertion_method="binary",
... )
>>> print(pipeline.report())
              precision    recall  f1-score   support

       False       0.85      0.68      0.75       256
        True       0.43      0.67      0.53        94

    accuracy                           0.68       350
   macro avg       0.64      0.67      0.64       350
weighted avg       0.74      0.68      0.69       350
>>> pipeline.predicted
                           sentence  label_keeping_everyone_together  prediction  correct
21    Okay so the circumference ...                           True         False    False
22                             Okay                          False         False     True
...                             ...                            ...           ...      ...
3035  Can we come back another time                           True          True     True
3042  You can talk about it befo...                          False          True    False

[350 rows x 4 columns]
\end{lstlisting}

\subsection{Models}
\label{sec:appendix package models}
We provide 102 pretrained SetFit classifiers as part of \texttt{EduBehaviors-Kit}.
We used the 50-session dataset of LLM-annotated assertions described in section \ref{sec:task} as training data for these classifiers. At the time of training, this dataset included 12,934 teacher utterances annotated by five LLM annotators. We used only the Claude Opus 5 and GPT 5.6 Luna annotations as training data, as this combination had the highest average pairwise inter-rater agreement across all assertions. We trained and evaluated a separate model for each assertion. For ease of use, we trained models only on utterances as-is, without context or speaker labels. Because assertions frequently have low base rates, we used a different stratified train and test split for each assertion. We split the data into $\frac{2}{3}$ training and $\frac{1}{3}$ test sets, grouping utterances, and then further split $\frac{1}{5}$ of the training set to use for evaluation during training. Stratification was done at the utterance-level. For each assertion, we trained a SetFit classifier using the SentenceTransformers \texttt{paraphrase-mpnet-base-v2} model \cite{reimers-2019-sentence-bert} as a base and the default SetFit \texttt{LogisticRegression} classification head. SetFit was selected after performing well in preliminary experiments when compared with other small encoder models, especially on assertions with high class imbalance. Models were trained using a learning rate of 2e-5, batch size of 16, 5,000 max steps, and 10 epochs.
In preliminary experiments, threshold tuning provided little to no improvement on model performance overall, and on multiple occasions decreased performance slightly, so the final \texttt{EduBehaviors-kit} annotation uses a simple argmax as the decision rule for classification.
Positive-class performance for each published assertion model is shown in table \ref{tab:encoders}. Note that evaluations happen on doubly-labeled data, so model disagreement indicated by lower $\alpha$ values bound precision/recall/F1 maximums. One model, for the assertion ``the utterance translates a prior student contribution into another language,'' yielded a test-set F1 score of 0, so it was not published.

{\footnotesize
\begin{longtable}{p{0.55\linewidth}rrrrr}
\caption{Test set Krippendorff's $\alpha$, precision, recall, F1 and base rate (\% positive) for each assertion encoder.}\label{tab:encoders} \\
\toprule
Assertion & $\alpha$ & Precision & Recall & F1 & Base Rate (\%) \\
\midrule
\endfirsthead
\toprule
Assertion & $\alpha$ & Precision & Recall & F1 & Base Rate (\%) \\
\midrule
\endhead
\midrule
\endfoot
\bottomrule
\endlastfoot
addresses another teacher & 0.65 & 0.25 & 0.14 & 0.18 & 0.5 \\
addresses the whole class & 0.71 & 0.79 & 0.79 & 0.79 & 57.9 \\
agrees with one or more students answer & 0.75 & 0.57 & 0.33 & 0.42 & 7.4 \\
asks a non mathematical question & 0.74 & 0.65 & 0.49 & 0.56 & 2.9 \\
asks a question to the whole class & 0.81 & 0.73 & 0.71 & 0.72 & 11.5 \\
asks one or more students for a connection between two mathematical representations & 0.38 & 0.67 & 0.33 & 0.44 & 1.3 \\
asks one or more students for an evaluation of a prior student contribution & 0.65 & 0.55 & 0.52 & 0.53 & 1.5 \\
asks one or more students for an explanation & 0.76 & 0.70 & 0.72 & 0.71 & 4.1 \\
asks one or more students for an explanation of how a method works & 0.73 & 0.68 & 0.59 & 0.63 & 2.1 \\
asks one or more students for clarification & 0.55 & 0.59 & 0.46 & 0.52 & 3.5 \\
asks one or more students for the definition of a math term & 0.58 & 0.75 & 0.35 & 0.47 & 0.3 \\
asks one or more students for the meaning of a mathematical symbol & 0.65 & 0.57 & 0.54 & 0.55 & 0.5 \\
asks one or more students for the next step in a procedure & 0.61 & 0.59 & 0.47 & 0.52 & 2.1 \\
asks one or more students for the purpose of a mathematical step & 0.73 & 0.66 & 0.62 & 0.64 & 0.7 \\
asks one or more students for the reasoning behind a prior student contribution & 0.78 & 0.57 & 0.52 & 0.55 & 1.4 \\
asks one or more students for the result of a math computation & 0.73 & 0.68 & 0.70 & 0.69 & 3.1 \\
asks one or more students for their reasoning & 0.71 & 0.73 & 0.72 & 0.72 & 3.5 \\
asks one or more students whether they agree & 0.71 & 0.66 & 0.53 & 0.59 & 1.6 \\
asks one or more students whether they agree with a prior student contribution & 0.67 & 0.71 & 0.56 & 0.63 & 1.0 \\
asks one or more students whether they are finished with a task & 0.48 & 0.45 & 0.35 & 0.39 & 0.3 \\
asks one or more students whether they are sure about an answer & 0.61 & 0.67 & 0.56 & 0.61 & 0.5 \\
asks one or more students whether they got a different answer than a prior student contribution & 0.37 & 0.37 & 0.56 & 0.44 & 0.2 \\
asks one or more students whether they heard a prior contribution & 0.91 & 0.75 & 0.86 & 0.80 & 0.1 \\
asks one or more students whether they understand something & 0.63 & 0.73 & 0.54 & 0.62 & 1.2 \\
contains a counting sequence & 0.79 & 0.81 & 0.81 & 0.81 & 1.3 \\
contains a greeting & 0.84 & 0.56 & 0.45 & 0.50 & 0.3 \\
contains a hypothetical scenario & 0.62 & 0.70 & 0.55 & 0.62 & 4.2 \\
contains a number & 0.90 & 0.94 & 0.96 & 0.95 & 27.9 \\
contains a personal stance & 0.72 & 0.74 & 0.67 & 0.70 & 7.5 \\
contains a politeness marker & 0.85 & 0.88 & 0.85 & 0.86 & 3.7 \\
contains a question & 0.88 & 0.92 & 0.90 & 0.91 & 24.1 \\
contains a real world reference pertaining to a topic outside of the classroom setting & 0.85 & 0.87 & 0.82 & 0.84 & 5.1 \\
contains a rhetorical question & 0.54 & 0.44 & 0.34 & 0.38 & 6.7 \\
contains a student s name & 0.96 & 0.95 & 0.98 & 0.97 & 8.3 \\
contains a time marker & 0.64 & 0.80 & 0.68 & 0.73 & 7.5 \\
contains a yes no question & 0.85 & 0.89 & 0.84 & 0.86 & 8.9 \\
contains affirmation & 0.68 & 0.76 & 0.67 & 0.71 & 8.9 \\
contains agreement & 0.66 & 0.71 & 0.55 & 0.62 & 7.4 \\
contains an acknowledgement & 0.70 & 0.67 & 0.55 & 0.61 & 13.8 \\
contains an apology & 0.87 & 0.89 & 0.81 & 0.85 & 0.4 \\
contains an explanation of a mathematical concept & 0.73 & 0.71 & 0.63 & 0.67 & 12.7 \\
contains an instruction & 0.75 & 0.78 & 0.77 & 0.77 & 20.1 \\
contains clarification & 0.51 & 0.68 & 0.59 & 0.64 & 22.9 \\
contains collaborative language & 0.67 & 0.80 & 0.81 & 0.81 & 17.3 \\
contains comparison terms & 0.66 & 0.76 & 0.70 & 0.73 & 7.0 \\
contains emotion & 0.68 & 0.76 & 0.70 & 0.73 & 5.0 \\
contains encouragement & 0.46 & 0.56 & 0.39 & 0.46 & 4.0 \\
contains fraction terms & 0.91 & 0.92 & 0.88 & 0.90 & 6.3 \\
contains gratitude & 0.65 & 0.93 & 0.76 & 0.84 & 1.8 \\
contains humor & 0.67 & 0.44 & 0.14 & 0.21 & 1.5 \\
contains informal language & 0.45 & 0.73 & 0.59 & 0.66 & 25.4 \\
contains math computation & 0.56 & 0.66 & 0.56 & 0.61 & 3.7 \\
contains math terms & 0.84 & 0.92 & 0.91 & 0.91 & 40.5 \\
contains measurement terms & 0.68 & 0.81 & 0.73 & 0.77 & 7.0 \\
contains negation & 0.90 & 0.95 & 0.95 & 0.95 & 8.4 \\
contains praise & 0.79 & 0.81 & 0.72 & 0.76 & 4.2 \\
contains self correction & 0.57 & 0.39 & 0.18 & 0.25 & 2.4 \\
contains the answer to a question & 0.64 & 0.52 & 0.33 & 0.40 & 10.4 \\
contains the final answer to a math problem & 0.64 & 0.53 & 0.40 & 0.46 & 4.8 \\
corrects a prior student contribution & 0.50 & 0.26 & 0.16 & 0.20 & 2.1 \\
disagrees with a prior contribution & 0.68 & 0.52 & 0.32 & 0.40 & 2.6 \\
evaluates a student response & 0.70 & 0.66 & 0.46 & 0.54 & 12.3 \\
grants permission & 0.74 & 0.59 & 0.47 & 0.53 & 1.9 \\
invites a student to speak next & 0.82 & 0.83 & 0.85 & 0.84 & 22.3 \\
invites an addition to a prior student contribution & 0.43 & 0.43 & 0.30 & 0.35 & 3.3 \\
invites participation & 0.76 & 0.84 & 0.83 & 0.84 & 34.3 \\
invites reading aloud & 0.84 & 0.85 & 0.72 & 0.78 & 0.4 \\
invites student thinking & 0.77 & 0.80 & 0.78 & 0.79 & 19.2 \\
invites writing on the board & 0.74 & 0.82 & 0.57 & 0.68 & 0.5 \\
invites writing on their own paper & 0.63 & 0.65 & 0.45 & 0.53 & 2.6 \\
is a declarative statement & 0.80 & 0.89 & 0.91 & 0.90 & 43.6 \\
is a one word agreement & 0.67 & 0.83 & 0.73 & 0.78 & 2.7 \\
is a single word & 0.94 & 0.96 & 0.98 & 0.97 & 12.0 \\
is a verbatim repetition of a prior student contribution & 0.54 & 0.43 & 0.16 & 0.23 & 2.7 \\
is an evaluative statement & 0.68 & 0.73 & 0.57 & 0.64 & 5.1 \\
is off topic & 0.61 & 0.50 & 0.18 & 0.27 & 1.4 \\
is unintelligible in context & 0.53 & 0.49 & 0.18 & 0.27 & 3.0 \\
is very short & 0.63 & 0.88 & 0.86 & 0.87 & 43.4 \\
manages classroom behavior & 0.61 & 0.75 & 0.58 & 0.65 & 11.3 \\
narrates steps being taken by the speaker & 0.50 & 0.64 & 0.48 & 0.55 & 6.4 \\
paraphrases a prior student contribution & 0.53 & 0.41 & 0.16 & 0.23 & 2.0 \\
proposes a problem solving strategy & 0.57 & 0.59 & 0.53 & 0.56 & 11.4 \\
reads a text aloud & 0.57 & 0.59 & 0.48 & 0.53 & 1.6 \\
references a prior lesson & 0.71 & 0.75 & 0.49 & 0.59 & 1.0 \\
references a procedure for completing the current task & 0.58 & 0.69 & 0.64 & 0.66 & 22.2 \\
references an activity from earlier in the session & 0.53 & 0.53 & 0.32 & 0.40 & 2.8 \\
references classroom materials & 0.70 & 0.82 & 0.69 & 0.75 & 6.6 \\
references logistics & 0.73 & 0.78 & 0.69 & 0.73 & 12.1 \\
references student behavior & 0.21 & 0.58 & 0.46 & 0.51 & 16.3 \\
references student work & 0.67 & 0.69 & 0.51 & 0.59 & 8.2 \\
repeats a number from a prior student contribution & 0.80 & 0.58 & 0.51 & 0.55 & 0.5 \\
repeats a phrase from a prior student contribution without adding new words & 0.51 & 0.53 & 0.23 & 0.32 & 3.5 \\
repeats a prior student contribution and adds new words & 0.56 & 0.36 & 0.14 & 0.20 & 3.2 \\
repeats a problem & 0.44 & 0.41 & 0.31 & 0.36 & 4.1 \\
repeats a single word from a prior student contribution & 0.60 & 0.53 & 0.41 & 0.46 & 3.6 \\
repeats a student answer & 0.71 & 0.54 & 0.24 & 0.33 & 4.8 \\
repeats prior speech & 0.63 & 0.46 & 0.25 & 0.32 & 12.1 \\
responds to a student contribution & 0.69 & 0.57 & 0.50 & 0.54 & 26.8 \\
states that the speaker is uncertain & 0.69 & 0.62 & 0.56 & 0.59 & 2.5 \\
states the goals of the lesson & 0.58 & 0.40 & 0.26 & 0.32 & 1.6 \\
summarizes a portion of the current lesson & 0.42 & 0.46 & 0.29 & 0.36 & 4.8 \\
transitions from one in class activity to another & 0.64 & 0.55 & 0.37 & 0.44 & 6.3 \\
\end{longtable}

}

\subsection{Encoders used in this paper}
\label{sec:appendix package encoders}
\begin{figure}[ht]
    \centering
    \includegraphics[width=0.8\columnwidth]{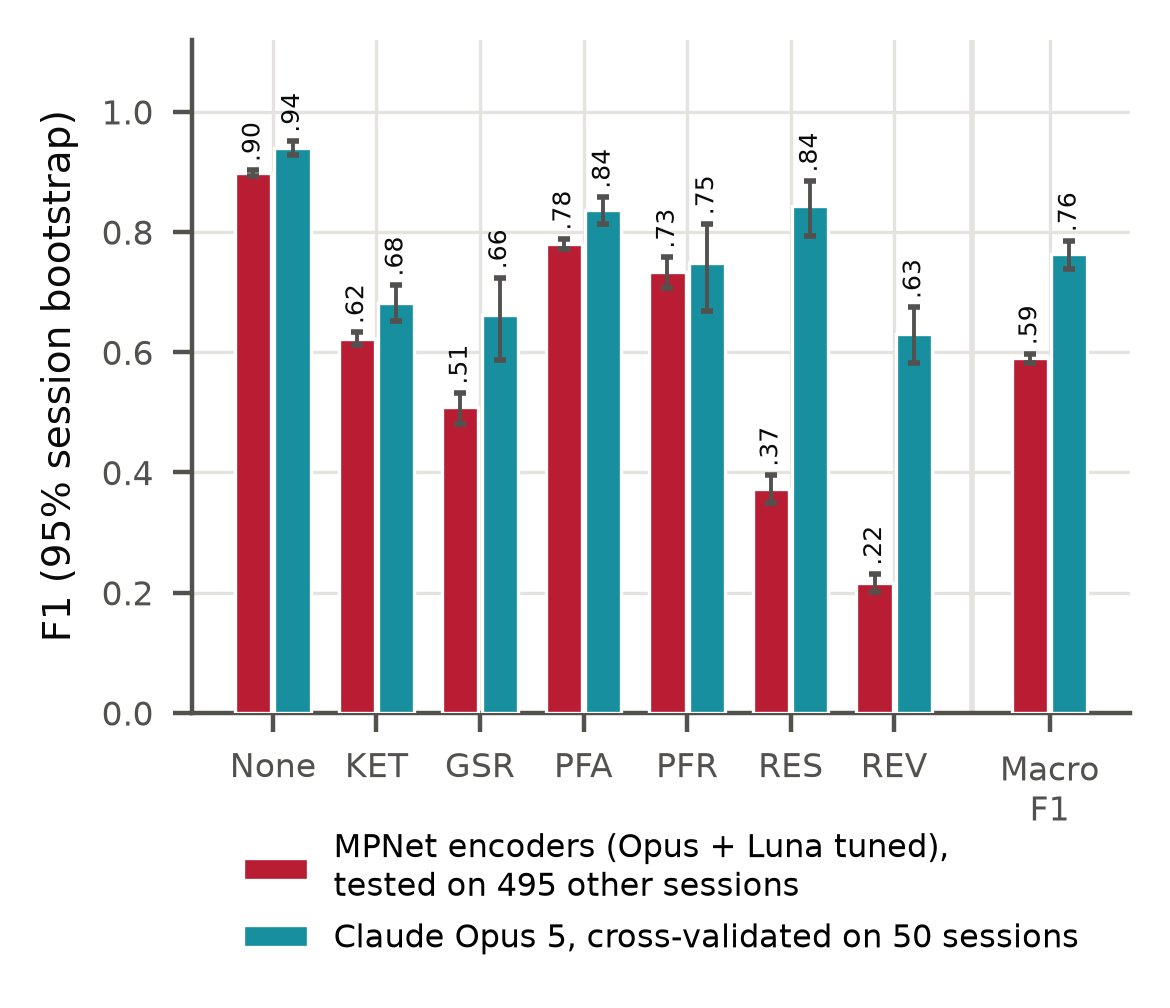}
    \caption{Per-move $F_1$ of the encoders (tested on the deployment sessions) against the best LLM annotator (cross-validated on the analysis sessions), with 95\% session-bootstrap intervals.}
    \label{fig:head-to-head}
\end{figure}

\begin{figure}[ht]
    \centering
    \includegraphics[width=0.8\columnwidth]{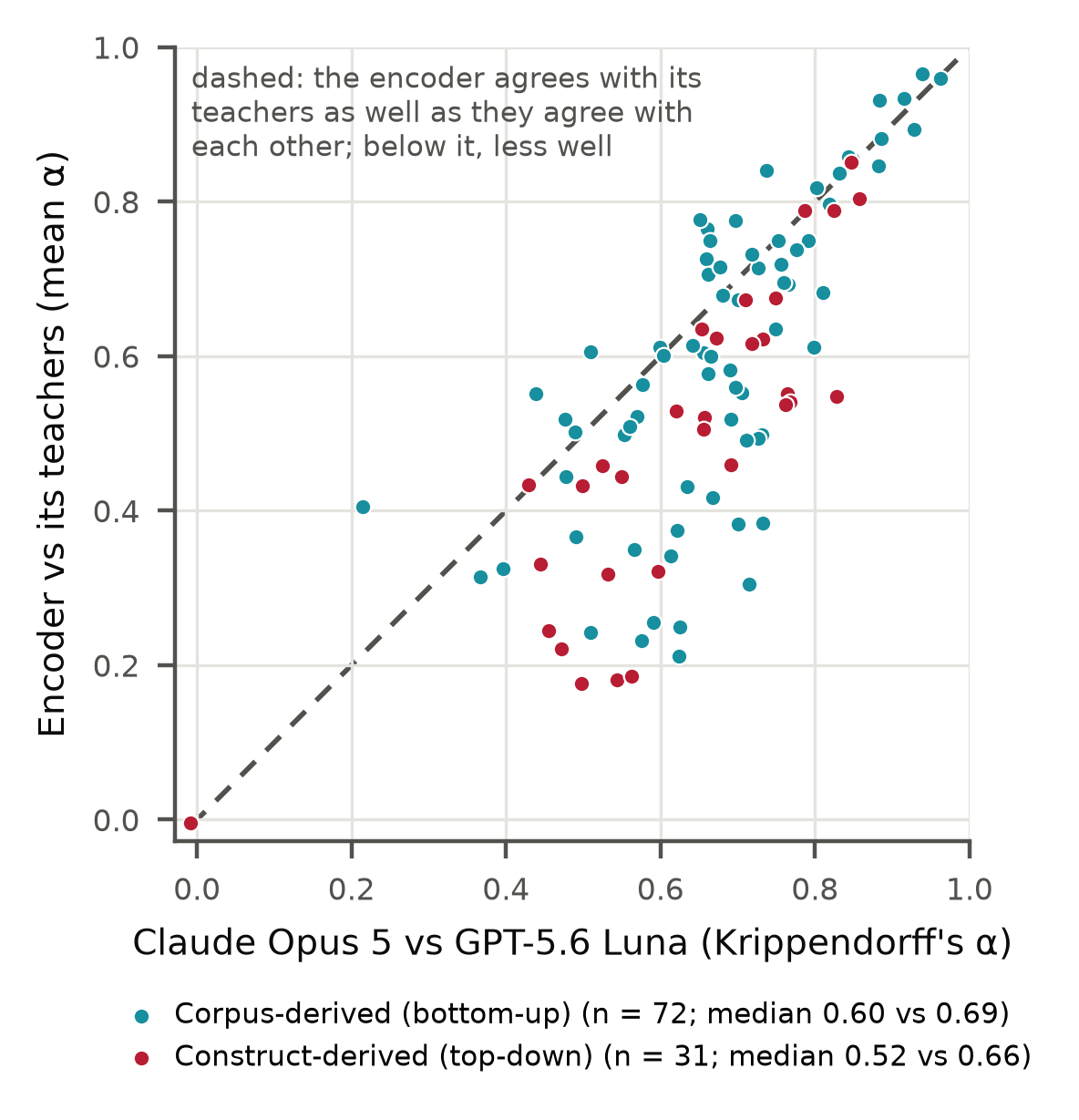}
    \caption{How well each assertion's encoder reproduces its teachers. Each point is one assertion, on held-out utterances: x is Krippendorff's $\alpha$ between the two teachers, Claude Opus 5 and GPT-5.6 Luna; y is the encoder's $\alpha$ with each teacher, averaged. Points below the dashed diagonal are assertions where the encoder agrees with its teachers less than they agree with each other.}
    \label{fig:encoder-agreement}
\end{figure}

In Table~\ref{tab:encoder-deployment} we show every encoder configuration: classifiers trained on the analysis sessions with each source of assertion labels, then applied to the deployment sessions using the encoders' predicted assertions. The bold row is the one reported in Table~\ref{tab:best-per-annotator}. In Figure~\ref{fig:head-to-head} we show per-move $F_1$ of the encoder pipeline on the deployment sessions next to the best LLM annotator's cross-validated $F_1$ on the analysis sessions. In Figure~\ref{fig:encoder-agreement} we show, for each assertion, how closely its encoder agrees with its two teachers compared with how closely the teachers agree with each other.

\begin{table*}[t]
\centering
\small
\begin{tabular}{llrr}
\toprule
Classifier trained on & Covariates & Macro-$F_1$ [95\% CI] & $\kappa$ \\
\midrule
\multicolumn{4}{l}{\textit{Trained on LLM labels}} \\
Claude Opus 5 labels & words + BU + TD & 0.578 [0.571, 0.585] & 0.629 \\
Claude Opus 5 labels & words + BU & 0.558 [0.550, 0.565] & 0.591 \\
GPT-5.6 Luna labels & words + BU + TD & 0.587 [0.579, 0.594] & 0.639 \\
GPT-5.6 Luna labels & words + BU & 0.559 [0.551, 0.567] & 0.599 \\
Claude + Luna labels, averaged & words + BU + TD & 0.587 [0.580, 0.594] & 0.634 \\
Claude + Luna labels, averaged & words + BU & 0.550 [0.543, 0.558] & 0.588 \\
Claude + Luna labels, stacked & words + BU + TD & \textbf{0.590 [0.583, 0.598]} & \textbf{0.636} \\
Claude + Luna labels, stacked & words + BU & 0.563 [0.555, 0.570] & 0.602 \\
\midrule
\multicolumn{4}{l}{\textit{Trained on the encoders' labels}} \\
the encoders' own labels & words + BU + TD & 0.584 [0.576, 0.591] & 0.640 \\
the encoders' own labels & words + BU & 0.557 [0.550, 0.564] & 0.615 \\
\midrule
\multicolumn{4}{l}{\textit{No assertions}} \\
words only & words & 0.388 [0.383, 0.394] & 0.366 \\
\bottomrule
\end{tabular}
\caption{Every encoder configuration. Classifiers are trained and tuned on the 50 analysis sessions and tested on 495 other sessions, whose assertions are the encoders' labels, each cut at a threshold tuned on held-out dev utterances. BU and TD are corpus- and construct-derived assertions. Bold marks the configuration reported in the main results table.}
\label{tab:encoder-deployment}
\end{table*}

\FloatBarrier

\section{Data}
\label{sec:appendix data}
In Tables~\ref{tab:data-subsets} and~\ref{tab:data-labels} we show how the TalkMoves corpus is split into the subsets this paper uses, and the gold talk moves in each. Every classifier is trained and scored on teacher utterances; the analysis sample is the 50 sessions that every LLM annotated. The vocabulary sessions are a disjoint set of 10 sessions used only to select words by frequency. The held-out sessions are a set of sessions used in an earlier iteration of assertion generation kept disjoint so as to not inflate cross-model agreement scores. The deployment sessions are the 496 sessions outside of the analysis set used to benchmark the performance of our encoder trained on assertion annotations. Note that one session in the TalkMoves dataset has no teacher utterances (Video Mosaic Grade 4 Building large models 3) and is thus catalogued here but excluded from the results page in the main paper. The TalkMoves dataset is distributed under a CC BY-NC-SA 4.0 license. We use it for non-commercial research consistent with these terms, and the assertion annotations we derive from it are released under the same CC BY-NC-SA 4.0 license.

\begin{table*}[t]
\centering
\small
\begin{tabular}{lrrrrr}
\toprule
Subset & Sessions & Utterances & Teacher & Student & Median teacher utt./session \\
\midrule
TalkMoves corpus & 566 & 234{,}002 & 173{,}846 & 58{,}926 & 298 \\
Analysis sample & 50 & 18{,}589 & 14{,}341 & 4{,}153 & 282 \\
Vocabulary sessions & 10 & 2{,}851 & 2{,}303 & 509 & 252 \\
Deployment sessions & 496 & 208{,}615 & 153{,}985 & 53{,}565 & 300.5 \\
Held out & 10 & 3{,}947 & 3{,}217 & 699 & 314.5 \\
\bottomrule
\end{tabular}
\caption{How the TalkMoves corpus is divided. The analysis sample is annotated by every LLM and carries every cross-validated result; word features are fitted on the vocabulary sessions only; the encoders are evaluated on the deployment sessions. Held-out sessions are used for none of these. Teacher utterances are the ones classified, each with a gold talk move.}
\label{tab:data-subsets}
\end{table*}

\begin{table*}[t]
\centering
\small
\begin{tabular}{lrrrrr}
\toprule
Gold talk move & TalkMoves corpus & Analysis sample & Vocabulary sessions & Deployment sessions & Held out \\
\midrule
None & 116{,}818 (67.2\%) & 9{,}943 (69.3\%) & 1{,}565 (68.0\%) & 102{,}989 (66.9\%) & 2{,}321 (72.1\%) \\
Keeping everyone together (KET) & 22{,}773 (13.1\%) & 1{,}783 (12.4\%) & 330 (14.3\%) & 20{,}319 (13.2\%) & 341 (10.6\%) \\
Getting students to relate (GSR) & 2{,}862 (1.6\%) & 240 (1.7\%) & 38 (1.7\%) & 2{,}551 (1.7\%) & 33 (1.0\%) \\
Pressing for accuracy (PFA) & 22{,}924 (13.2\%) & 1{,}733 (12.1\%) & 275 (11.9\%) & 20{,}519 (13.3\%) & 397 (12.3\%) \\
Pressing for reasoning (PFR) & 2{,}039 (1.2\%) & 150 (1.0\%) & 25 (1.1\%) & 1{,}837 (1.2\%) & 27 (0.8\%) \\
Restating (RES) & 2{,}437 (1.4\%) & 186 (1.3\%) & 23 (1.0\%) & 2{,}196 (1.4\%) & 32 (1.0\%) \\
Revoicing (REV) & 3{,}993 (2.3\%) & 306 (2.1\%) & 47 (2.0\%) & 3{,}574 (2.3\%) & 66 (2.1\%) \\
\midrule
Teacher utterances & 173{,}846 & 14{,}341 & 2{,}303 & 153{,}985 & 3{,}217 \\
\bottomrule
\end{tabular}
\caption{Gold talk moves of the teacher utterances in each subset: count and share of the subset's teacher utterances.}
\label{tab:data-labels}
\end{table*}

\clearpage

\section{Prompts}
\label{sec:appendix prompts}

\subsection{Annotation}
\label{sec:appendix prompts annotation}
Every annotator received the same prompt, once per session and assertion. The system message is below; \texttt{\{statement\}} is the assertion. Definitions of specific terms in assertion text (e.g. question, answer, student contribution) are defined once with definitions repeated across any assertion using them and included in the definitions block.

\begin{lstlisting}[title={System message}]
<persona>
You are an expert annotation assistant for educational research. Your goal is to read the ENTIRE transcript of a mathematics lesson and identify every utterance where a given assertion is true.
</persona>
<context>
Each lesson is around 55 minutes on average and you will be given its full transcript, not an excerpt. Each line begins with its utterance index, then a speaker in brackets, either [Teacher] or [Student]. Use the whole transcript to understand what's happening across the lesson.
</context>
<assertion_definition>
An assertion is a statement about an utterance that can be unambiguously resolved to either True or False. Think about it as a Yes/No question that you can go through the transcript with and apply to every utterance to get a (true/false) value. 
</assertion_definition>
<task>
Your goal is to read the entire transcript and identify every TEACHER utterance where the assertion below is true. Only teacher utterances are eligible as answers — never include a student utterance's index, even if the assertion's wording could plausibly apply to one.
</task>
<assertion>
Assertion: "{statement}"
</assertion>
<conventions>
The <definitions> section below provides an unambiguous definition of the assertion, broken into definitional components. The supplied definition is the definition you must apply to each utterance; do not substitute any other reading of the assertion's wording.
Each component is listed as: "<component text>" (<component type>), definition: <definition>. The component types are:
- Verb frame: the general form of the assertion, with a placeholder where a more specific component goes. The placeholder is X (filled by a noun phrase or a clause), NOUN (filled by a noun phrase), ADJECTIVE (filled by an adjective), or VERB_PHRASE (filled by a verb phrase).
- Noun phrase: the thing, event, or kind of language that fills X or NOUN.
- Clause: a short statement that fills X in a verb frame of the form "... whether X".
- Adjective: a property of the whole utterance that fills ADJECTIVE.
- Verb phrase: the action or relationship that fills VERB_PHRASE; its definition is already the complete definition of the assertion.
To get the full meaning of the assertion, take the verb frame's definition and replace its placeholder with the filling component's definition.
Include an utterance only if the definition clearly applies to it; if it applies only partially or arguably, leave it out.
In these definitions, "the focal utterance" is the teacher utterance whose index you are deciding whether to include.
The assertion is restated after the definitions.
</conventions>
<definitions>
{definitions}
</definitions>
<assertion>
Assertion: "{statement}"
</assertion>
<instructions>
- Consider every TEACHER utterance in the transcript as a candidate; student utterances are context only, never candidates.
- Read the assertion text as literally as possible.
- An utterance's index belongs in your answer if and only if the assertion is true for that specific utterance.
- For each utterance you include, also report your confidence (1-5) that the assertion is true for it: 5 = no doubt, 1 = essentially a coin flip.
- It is common and expected for the assertion to be true for zero, one, or many utterances in the same session.
</instructions>
<output_format>
- Respond with ONLY a single JSON object of the exact form {"true_indices": [{"idx": <int>, "confidence": <int>}, ...]}, one entry per teacher utterance index where the assertion is true, in ascending order by idx.
- If the assertion is true for no utterance in this session, respond with {"true_indices": []}.
- Respond with ONLY that JSON object and nothing else — no markdown code fences, no justification, no explanation, and no text before or after it.
</output_format>
\end{lstlisting}

\begin{lstlisting}[title={User message}]
<transcript>
The following is the COMPLETE transcript of one classroom conversation during a mathematics lesson, start to finish.

{transcript}
</transcript>
\end{lstlisting}
\texttt{\{transcript\}} is the whole session, one line per utterance in the form \texttt{<index>: [Teacher] <text>} or \texttt{<index>: [Student] <text>}.

\subsection{Construct-derived assertion generation}
\label{sec:appendix prompts top-down}
The construct-derived assertions were generated by Gemini 3.1 Pro at temperature 0 from this prompt. \texttt{\{bottom\_up\}} lists the 72 corpus-derived assertions; \texttt{\{frames\}} lists the verb frames used in their definitions; \texttt{\{codebook\}} is the Teacher TalkMoves section of the TalkMoves coding manual.

\begin{lstlisting}[title={System message}]
<persona>
You are an expert interpreter of codebooks for qualitative coding of classroom talk. You read a coding manual that assigns labels to individual teacher utterances and write assertions: short statements a coder can judge TRUE or FALSE for one utterance.
</persona>
<assertion_rules>
(1) An assertion is TRUE or FALSE for one teacher utterance, which is read together with the utterances that come before it.
(2) Every statement begins "The utterance".
(3) One predicate per assertion. Several subjects may be joined ("one or more students"); two actions or behaviors may not be joined with "or" or "and". If you need a semicolon, write two assertions.
(4) Code what the utterance DOES, not its wording. If the same act said in completely different words would change the judgment, the assertion is invalid. Never match specific words, phrases, names, or punctuation, and never quote the manual's example utterances.
(5) No parentheticals and no carve-outs: no "unless", "except", "other than", "not including", and no "(e.g. ...)". Whatever precision the statement needs goes in its definition.
(6) A definition says what the assertion means. It never mentions, depends on, or excludes any other assertion.
</assertion_rules>
<definitions_format>
Every assertion is a verb frame filled by one component. The verb frames already in use, with their definitions:
{frames}
Reuse one of these frames whenever it fits. If none fits, you may write a new frame and must define it in the same style. The filling component is one of: Noun phrase (fills X or NOUN), Clause (fills X in "... whether X"), Adjective (fills ADJECTIVE), or Verb phrase (fills VERB_PHRASE; its definition is the complete definition of the assertion). Write each definition so that two coders reading only the definition would judge the same utterance the same way. Define the core of the behavior: do not narrow a definition with qualifiers such as exact, verbatim, identical, specific or explicit unless the manual itself requires that restriction. A definition that excludes the ordinary forms of a behavior makes the assertion too rare to be useful. In definitions, "the focal utterance" means the teacher utterance being judged.
</definitions_format>
<bottom_up_assertions>
These assertions are already in use. Each is shown with its definition.
{bottom_up}
</bottom_up_assertions>
<task>
For each label in the manual below, including the residual None label, write assertions that:
(a) are evidence for that label: an utterance where the assertion is TRUE should be more likely to carry that label;
(b) are well-defined under the rules and format above;
(c) are not restatements of a bottom-up assertion. Overlapping a bottom-up assertion is fine when the new assertion is more specific evidence for this label; an assertion with the same meaning as a bottom-up assertion is not;
(d) are TRUE for at least three teacher utterances in a typical 55-minute math lesson. Aim first at the ordinary, everyday forms in which each label shows up, not its rarest forms. For each assertion, estimate how many teacher utterances in a typical lesson it would be TRUE for, and do not include any you estimate at fewer than three.
Write 4 to 6 assertions per label, and up to 8 for labels that apply to many teacher utterances in an ordinary lesson. A label may have fewer only if the bottom-up assertions already cover nearly all of it.
</task>
<output_format>
A single JSON object. Each key is a label from the manual; each value is an array of:
{"statement": "The utterance ...",
 "frame": "<verb frame text>",
 "frame_definition": "<only if the frame is new, else omit>",
 "component": "<the filling text>",
 "component_type": "Noun phrase | Clause | Adjective | Verb phrase",
 "component_definition": "...",
 "not_covered_because": "<one sentence>",
 "times_per_lesson": <your estimate from (d), an integer>}
</output_format>

\end{lstlisting}

\begin{lstlisting}[title={User message}]
<codebook>
{codebook}</codebook>
Now answer:
\end{lstlisting}

\clearpage
\section{Agreement}
\label{sec:appendix agreement}
\begin{figure}[ht]
    \centering
    \includegraphics[width=0.8\columnwidth]{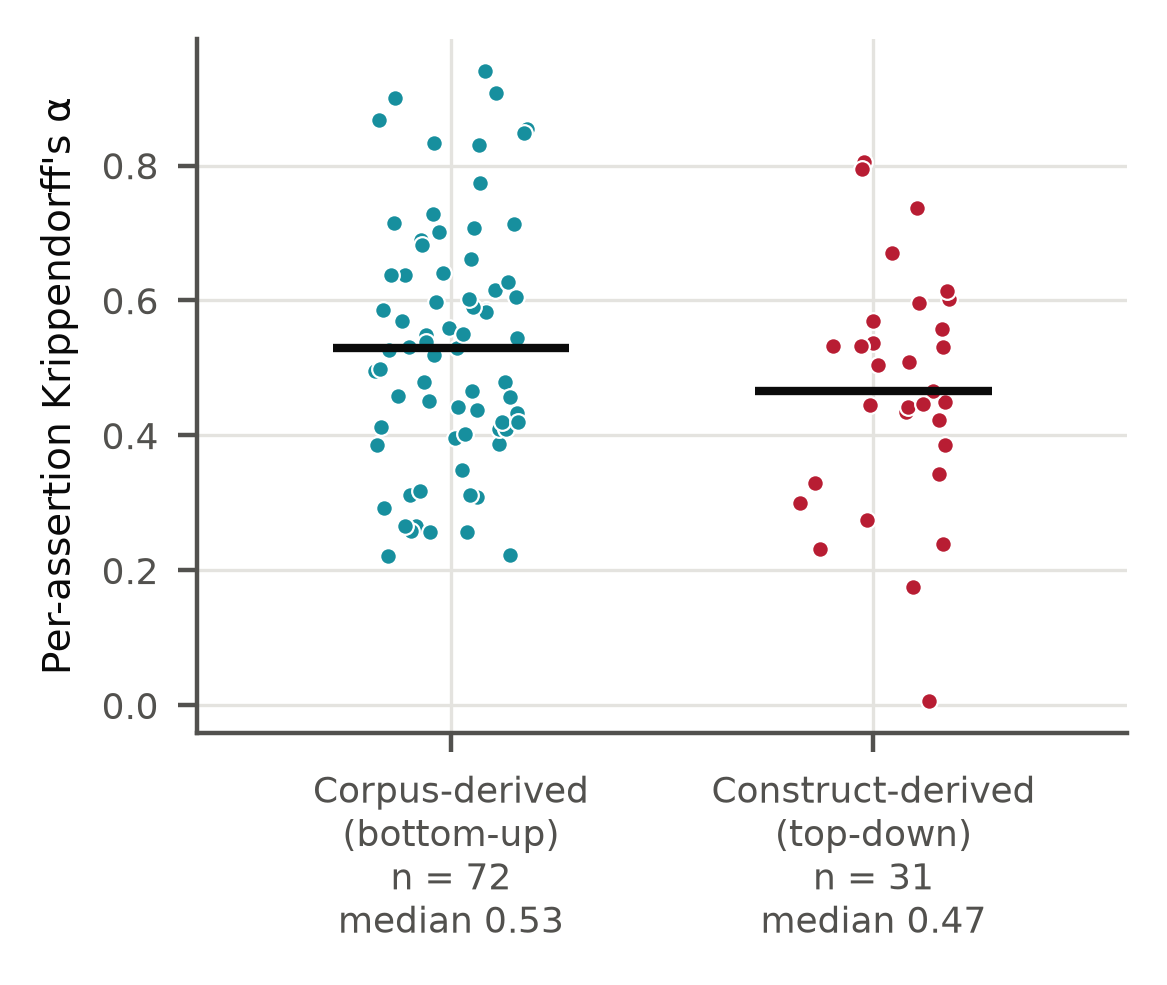}
    \caption{Cross-annotator agreement of each assertion (Krippendorff's $\alpha$ over the five annotators), split by source. Each point is one assertion, jittered horizontally; the black bar is the source's median.}
    \label{fig:agreement-by-source}
\end{figure}

\begin{figure}[ht]
    \centering
    \includegraphics[width=0.8\columnwidth]{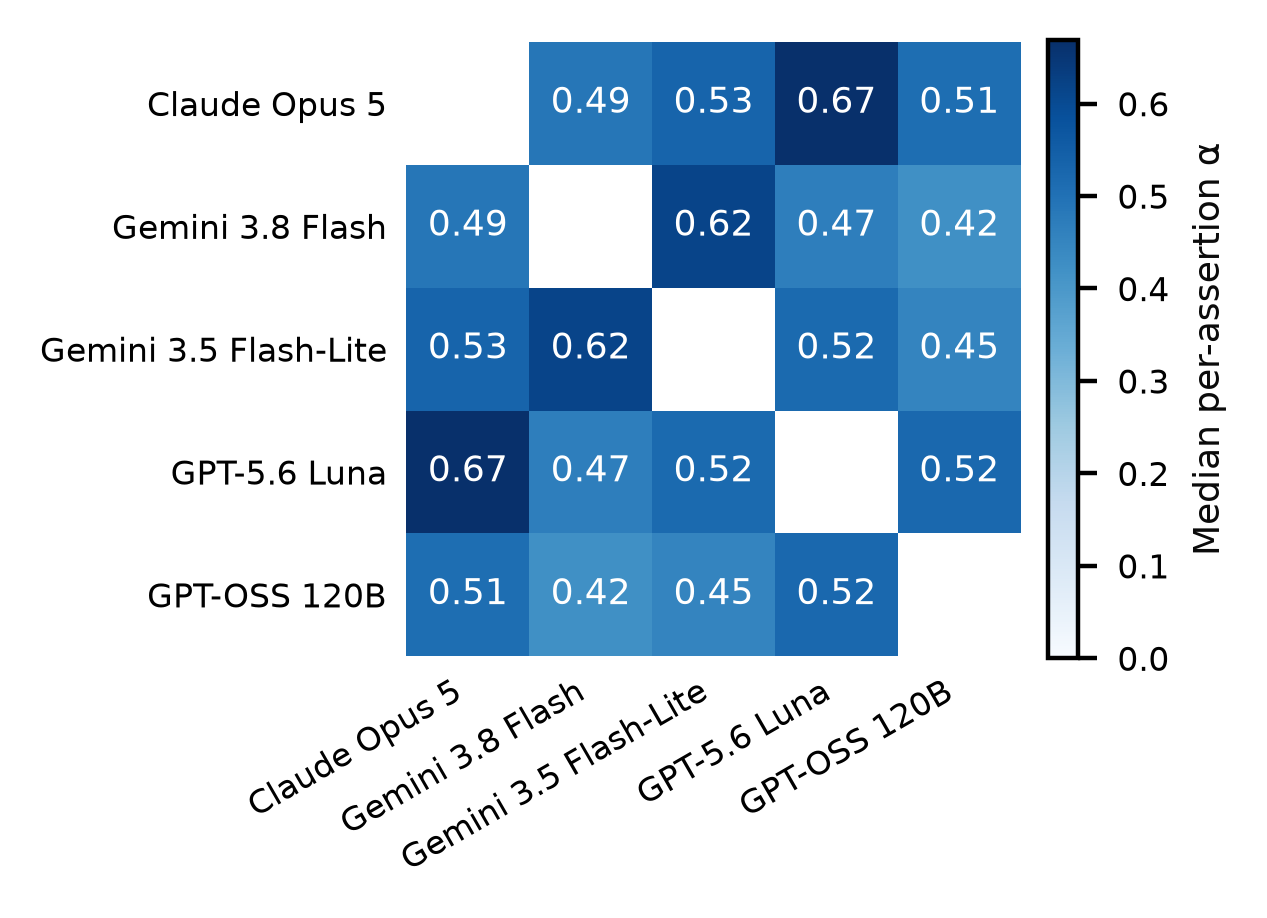}
    \caption{Agreement between each pair of annotators: for every pair, the median over assertions of the pair's Krippendorff's $\alpha$ on that assertion.}
    \label{fig:pairwise-agreement}
\end{figure}

In Figure~\ref{fig:agreement-by-source} we show how much the five annotators agree on each assertion, separately for corpus-derived and construct-derived assertions. We see higher agreement in the corpus-derived set, but significant variability in both. In Figure~\ref{fig:pairwise-agreement} we show agreement for each pair of annotators, as the median $\alpha$ over assertions. Our top agreeing pair is Claude Opus 5 and GPT 5.6 Luna, while our least agreeing pair were Gemini 3.8 Flash and GPT-OSS 120B.
\par
\begin{figure}[!htbp]
    \centering
    \includegraphics[width=0.8\columnwidth]{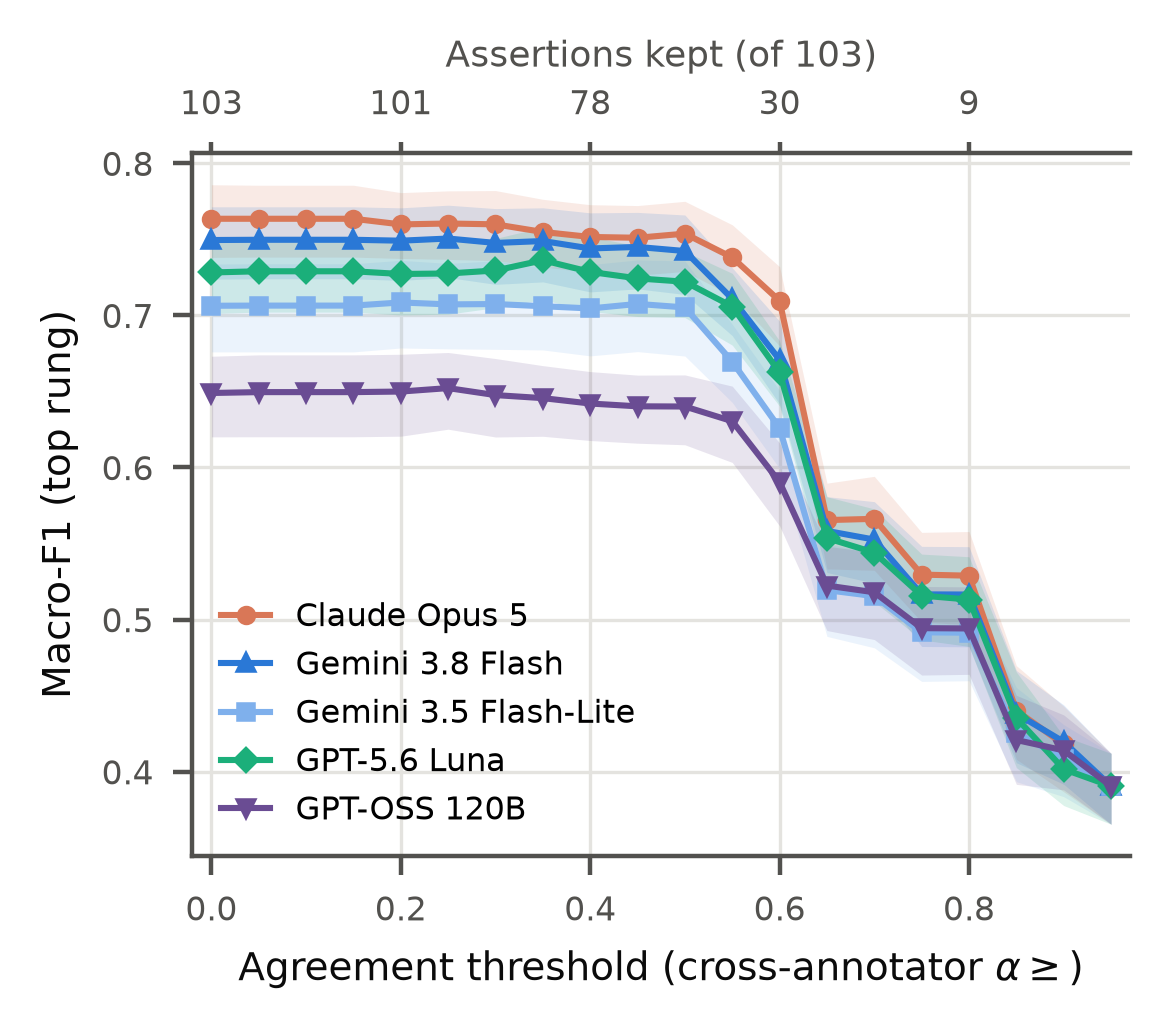}
    \caption{Macro-$F_1$ of each annotator's predictions using all assertions with agreement exceeding the threshold against the threshold, in steps of 0.05. The top axis gives the number of assertions kept; above $\alpha = 0.94$ none remain and every annotator falls to words only. Bands are 95\% session-bootstrap intervals.}
    \label{fig:agreement-sweep}
\end{figure}

 In Figure~\ref{fig:agreement-sweep} we show the effect of restricting the set of assertions to only those with a sufficiently high cross-model agreement with respect to the final predictions' macro-F1. Performance is relatively flat up to around a restriction of $\alpha \geq 0.5$, then drops harshly as the number of retained assertions falls to zero.

\clearpage
\section{Model Inference and Costs}
\label{sec:appendix costs}
In Table~\ref{tab:cost} we show what annotating the analysis sample cost for each annotator. One call annotates one assertion over one whole session, so each annotator made one call per session and assertion. All models were queried at their respective ``medium" reasoning effort configuration. For reproducibility's sake, models were run at temperature 0 when possible, although Claude Opus 5 and GPT 5.6 Luna had restrictions on temperature for our endpoint, with those models running at temperature 1.0. Otherwise, no decoding method is specified beyond provider default. All models were queried through Gemini Agent Platform except GPT 5.6 Luna which was queried through a university-specific API.
\par
\begin{figure}[ht]
    \centering
    \includegraphics[width=0.8\columnwidth]{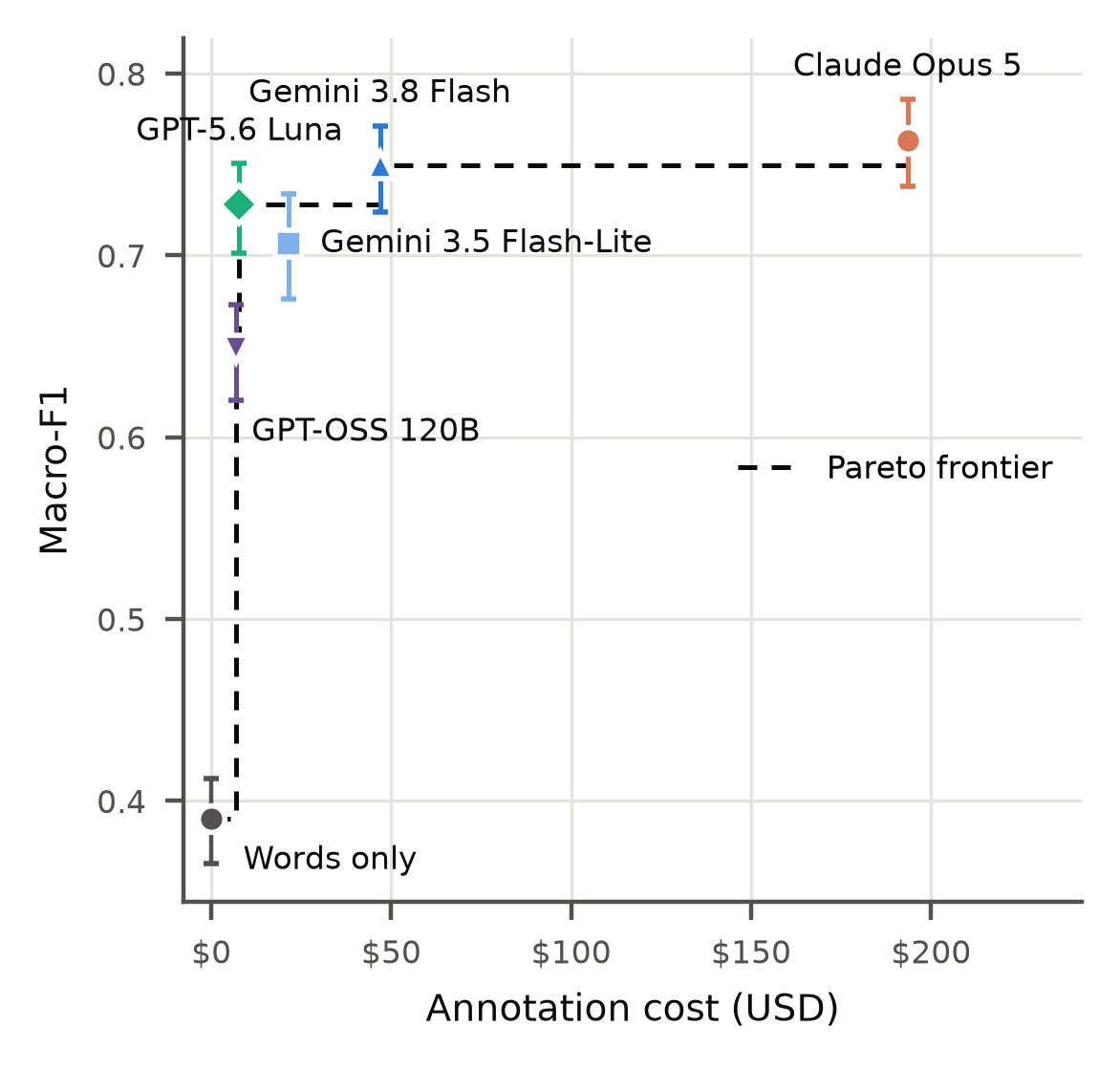}
    \caption{Annotation cost over the 50 analysis sessions against macro-$F_1$, one point per annotator, with 95\% session-bootstrap intervals. Words only, which needs no annotation, sits at \$0.}
    \label{fig:cost-performance}
\end{figure}

 In Figure~\ref{fig:cost-performance} we show each annotator's cost against its macro-$F_1$ in the all assertion configuration and trace the Pareto-frontier curve: at each price point, the curve points to the best-performing annotator, with all annotators lying on the curve except for Gemini 3.5 Flash Lite.

\begin{table*}[t]
\centering
\small
\begin{tabular}{llrrrrrrrr}
\toprule
 & & & \multicolumn{2}{c}{Tokens per call} & \multicolumn{2}{c}{USD per million tokens} & \multicolumn{3}{c}{Cost (USD)} \\
\cmidrule(lr){4-5} \cmidrule(lr){6-7} \cmidrule(lr){8-10}
 & & & & & & & & \multicolumn{2}{c}{Per session} \\
\cmidrule(lr){9-10}
Annotator & Mode & Calls & Input & Output & Input & Output & Total & One assertion & All 103 \\
\midrule
Claude Opus 5 & batch & 5{,}150 & 8{,}815 & 959 & 2.750 & 13.750 & 193.49 & 0.0376 & 3.87 \\
Gemini 3.8 Flash & batch & 5{,}150 & 6{,}897 & 3{,}462 & 0.375 & 1.875 & 46.83 & 0.0091 & 0.94 \\
Gemini 3.5 Flash-Lite & batch & 5{,}150 & 6{,}901 & 2{,}494 & 0.150 & 1.250 & 21.44 & 0.0042 & 0.43 \\
GPT-5.6 Luna & sync & 5{,}150 & 6{,}199 & 1{,}484 & 0.100 & 0.600 & 7.78 & 0.0015 & 0.16 \\
GPT-OSS 120B & batch & 5{,}150 & 6{,}427 & 6{,}314 & 0.040 & 0.160 & 6.74 & 0.0013 & 0.13 \\
\midrule
Total & & 25{,}750 & & & & & 276.29 & 0.0536 & 5.53 \\
\bottomrule
\end{tabular}
\caption{Annotation cost per annotator over the 50 analysis sessions: one call per session and assertion, billed at each provider's rates for the mode used (batch where available). Per session: one assertion is the mean cost of a call; all 103 is the total over the 50 sessions. The total row is the whole panel.}
\label{tab:cost}
\end{table*}

\FloatBarrier
\raggedbottom

\clearpage
\section{List of Assertions}
\label{sec:appendix assertions}
 In Table~\ref{tab:assertions} we list every assertion $A$ by source, with its cross-annotator agreement as measured by Krippendorff's $\alpha$ across the five annotator panel along with fire rate as measured by the fraction of teacher utterances $u$ with $A(u) = 1$

\begin{longtable}{rp{0.62\linewidth}rr}
\caption{Every assertion, by source, with its cross-annotator agreement (Krippendorff's $\alpha$; -- where no annotator ever judged it true) and mean fire rate across the five annotators.}\label{tab:assertions} \\
\toprule
\# & Assertion & $\alpha$ & Fire rate \\
\midrule
\endfirsthead
\toprule
\# & Assertion & $\alpha$ & Fire rate \\
\midrule
\endhead
\bottomrule
\endfoot
\multicolumn{4}{l}{\textit{Corpus-derived (bottom-up), 72 assertions}} \\
1 & The utterance addresses the whole class & 0.56 & 62.4\% \\
2 & The utterance agrees with one or more students' answer & 0.46 & 4.1\% \\
3 & The utterance asks a question to the whole class & 0.68 & 10.8\% \\
4 & The utterance asks one or more students for an explanation & 0.59 & 3.3\% \\
5 & The utterance asks one or more students for clarification & 0.31 & 2.9\% \\
6 & The utterance asks one or more students for their reasoning & 0.60 & 2.9\% \\
7 & The utterance asks one or more students whether they agree & 0.53 & 1.0\% \\
8 & The utterance asks one or more students whether they are sure about an answer & 0.27 & 0.3\% \\
9 & The utterance asks one or more students whether they understand something & 0.55 & 0.8\% \\
10 & The utterance contains a counting sequence & 0.71 & 1.1\% \\
11 & The utterance contains a greeting & 0.70 & 0.3\% \\
12 & The utterance contains a hypothetical scenario & 0.44 & 2.7\% \\
13 & The utterance contains a number & 0.87 & 26.4\% \\
14 & The utterance contains a personal stance & 0.48 & 5.5\% \\
15 & The utterance contains a politeness marker & 0.83 & 3.1\% \\
16 & The utterance contains a question & 0.85 & 21.4\% \\
17 & The utterance contains a real-world reference pertaining to a topic outside of the classroom setting & 0.71 & 4.4\% \\
18 & The utterance contains a rhetorical question & 0.39 & 5.0\% \\
19 & The utterance contains a student's name & 0.94 & 8.6\% \\
20 & The utterance contains a time marker & 0.58 & 6.1\% \\
21 & The utterance contains affirmation & 0.54 & 6.7\% \\
22 & The utterance contains agreement & 0.50 & 5.3\% \\
23 & The utterance contains an acknowledgement & 0.57 & 10.7\% \\
24 & The utterance contains an apology & 0.83 & 0.4\% \\
25 & The utterance contains an explanation of a mathematical concept & 0.41 & 8.3\% \\
26 & The utterance contains an instruction & 0.60 & 15.7\% \\
27 & The utterance contains clarification & 0.26 & 11.9\% \\
28 & The utterance contains collaborative language & 0.69 & 15.0\% \\
29 & The utterance contains comparison terms & 0.59 & 5.6\% \\
30 & The utterance contains emotion & 0.41 & 3.4\% \\
31 & The utterance contains encouragement & 0.31 & 2.4\% \\
32 & The utterance contains fraction terms & 0.85 & 6.2\% \\
33 & The utterance contains gratitude & 0.77 & 1.4\% \\
34 & The utterance contains humor & 0.50 & 1.1\% \\
35 & The utterance contains informal language & 0.27 & 18.4\% \\
36 & The utterance contains math computation & 0.53 & 2.5\% \\
37 & The utterance contains math terms & 0.73 & 34.9\% \\
38 & The utterance contains measurement terms & 0.59 & 5.2\% \\
39 & The utterance contains negation & 0.90 & 8.4\% \\
40 & The utterance contains praise & 0.72 & 3.2\% \\
41 & The utterance contains self-correction & 0.41 & 1.8\% \\
42 & The utterance contains the answer to a question & 0.39 & 6.0\% \\
43 & The utterance contains the final answer to a math problem & 0.47 & 2.8\% \\
44 & The utterance disagrees with a prior contribution & 0.43 & 1.5\% \\
45 & The utterance evaluates a student response & 0.52 & 7.8\% \\
46 & The utterance grants permission & 0.55 & 1.2\% \\
47 & The utterance invites participation & 0.66 & 27.4\% \\
48 & The utterance invites student thinking & 0.61 & 15.4\% \\
49 & The utterance is a declarative statement & 0.64 & 39.6\% \\
50 & The utterance is a one-word agreement & 0.46 & 2.7\% \\
51 & The utterance is a single word & 0.91 & 11.7\% \\
52 & The utterance is off-topic & 0.44 & 1.3\% \\
53 & The utterance is unintelligible in context & 0.32 & 2.1\% \\
54 & The utterance is very short & 0.64 & 34.9\% \\
55 & The utterance manages classroom behavior & 0.45 & 7.7\% \\
56 & The utterance narrates steps being taken by the speaker & 0.35 & 4.1\% \\
57 & The utterance proposes a problem-solving strategy & 0.31 & 6.4\% \\
58 & The utterance reads a text aloud & 0.48 & 1.3\% \\
59 & The utterance references a prior lesson & 0.62 & 0.6\% \\
60 & The utterance references a procedure for completing the current task & 0.29 & 12.6\% \\
61 & The utterance references an activity from earlier in the session & 0.26 & 2.2\% \\
62 & The utterance references classroom materials & 0.63 & 4.9\% \\
63 & The utterance references logistics & 0.54 & 7.5\% \\
64 & The utterance references student behavior & 0.22 & 9.7\% \\
65 & The utterance references student work & 0.42 & 5.7\% \\
66 & The utterance repeats a problem & 0.26 & 2.2\% \\
67 & The utterance repeats a student answer & 0.64 & 3.6\% \\
68 & The utterance repeats prior speech & 0.42 & 7.3\% \\
69 & The utterance responds to a student contribution & 0.53 & 18.9\% \\
70 & The utterance states that the speaker is uncertain & 0.40 & 1.7\% \\
71 & The utterance summarizes a portion of the current lesson & 0.22 & 2.5\% \\
72 & The utterance transitions from one in-class activity to another & 0.40 & 3.9\% \\
\midrule
\multicolumn{4}{l}{\textit{Construct-derived (top-down), 31 assertions}} \\
73 & The utterance addresses another teacher & 0.45 & 0.5\% \\
74 & The utterance asks a non-mathematical question & 0.60 & 2.6\% \\
75 & The utterance asks one or more students for a connection between two mathematical representations & 0.24 & 0.7\% \\
76 & The utterance asks one or more students for an evaluation of a prior student contribution & 0.56 & 1.0\% \\
77 & The utterance asks one or more students for an explanation of how a method works & 0.45 & 1.3\% \\
78 & The utterance asks one or more students for the definition of a math term & 0.42 & 0.1\% \\
79 & The utterance asks one or more students for the meaning of a mathematical symbol & 0.33 & 0.2\% \\
80 & The utterance asks one or more students for the next step in a procedure & 0.43 & 1.3\% \\
81 & The utterance asks one or more students for the purpose of a mathematical step & 0.23 & 1.2\% \\
82 & The utterance asks one or more students for the reasoning behind a prior student contribution & 0.61 & 1.1\% \\
83 & The utterance asks one or more students for the result of a math computation & 0.51 & 2.1\% \\
84 & The utterance asks one or more students whether they agree with a prior student contribution & 0.60 & 0.5\% \\
85 & The utterance asks one or more students whether they are finished with a task & 0.54 & 0.3\% \\
86 & The utterance asks one or more students whether they got a different answer than a prior student contribution & 0.18 & 0.1\% \\
87 & The utterance asks one or more students whether they heard a prior contribution & 0.67 & 0.1\% \\
88 & The utterance contains a yes/no question & 0.80 & 8.3\% \\
89 & The utterance corrects a prior student contribution & 0.30 & 1.0\% \\
90 & The utterance invites a student to speak next & 0.53 & 16.0\% \\
91 & The utterance invites an addition to a prior student contribution & 0.28 & 1.8\% \\
92 & The utterance invites reading aloud & 0.80 & 0.3\% \\
93 & The utterance invites writing on the board & 0.44 & 0.3\% \\
94 & The utterance invites writing on their own paper & 0.45 & 1.6\% \\
95 & The utterance is a verbatim repetition of a prior student contribution & 0.57 & 1.7\% \\
96 & The utterance is an evaluative statement & 0.53 & 4.0\% \\
97 & The utterance paraphrases a prior student contribution & 0.34 & 1.3\% \\
98 & The utterance repeats a number from a prior student contribution & 0.74 & 0.6\% \\
99 & The utterance repeats a phrase from a prior student contribution without adding new words & 0.53 & 2.3\% \\
100 & The utterance repeats a prior student contribution and adds new words & 0.39 & 2.0\% \\
101 & The utterance repeats a single word from a prior student contribution & 0.50 & 2.6\% \\
102 & The utterance states the goals of the lesson & 0.47 & 1.0\% \\
103 & The utterance translates a prior student contribution into another language & 0.01 & 0.3\% \\

\end{longtable}

\clearpage
\section{Classification}
\label{sec:appendix classification}
\begin{figure*}[!tp]
    \centering
    \includegraphics[width=0.9\textwidth]{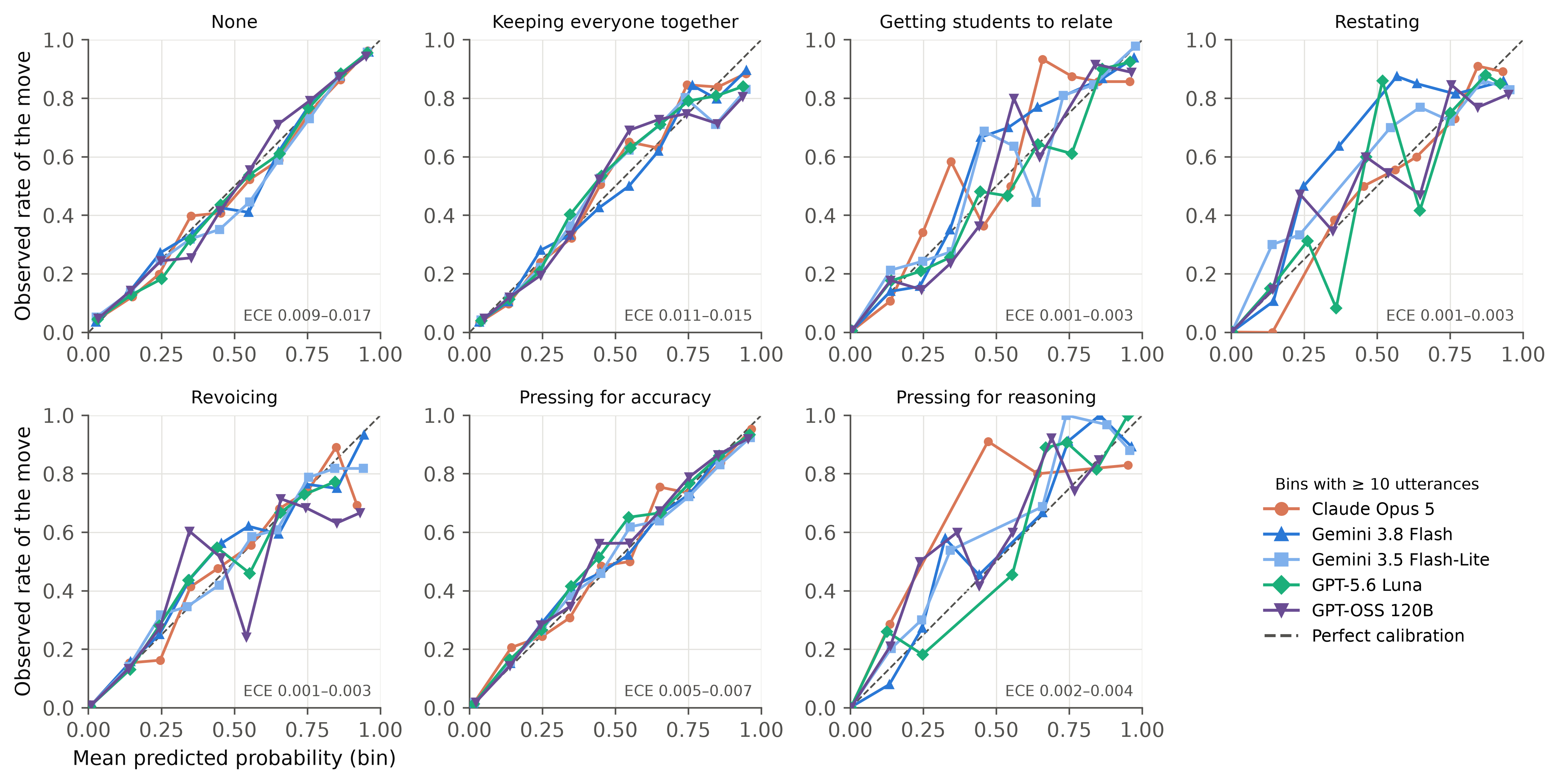}
    \caption{Calibration of each annotator's classifier, per move: held-out utterances are put in ten equal-width bins by predicted probability, and each bin's mean probability is plotted against how often the move is actually the gold label. Only bins with at least ten utterances are drawn. The dashed diagonal is perfect calibration; each panel gives the range of expected calibration error (ECE) across annotators.}
    \label{fig:reliability}
\end{figure*}

In Table~\ref{tab:fit-grid} we show every configuration we fitted: for each annotator, we show the scores with and without the construct-derived assertions. In addition to the full configurations for each annotator, we also consider three combined annotator configurations where an assertion value for a considered true if: (1) all annotators returned true for that assertion, (2) any annotator returned true for that assertion, or (3) the majority of annotators returned true for that assertion. The results for our annotators expands on that included in the main paper, and for the combined annotators we actually see our highest Macro-F1 across all configurations, although its gain over Claude Opus 5 is practically insignificant. In Figure~\ref{fig:reliability} we show whether the classifiers' predicted probabilities are calibrated: for utterances assigned a given likelihood of being a specific gold label, we plot the fraction that did indeed have that gold label. In doing so, we see that calibration varies by annotator and label, with some classes approximately calibrated (PFA) while others show greater divergences (PFR).

\par
\begin{figure}[t]
    \centering
    \includegraphics[width=\columnwidth]{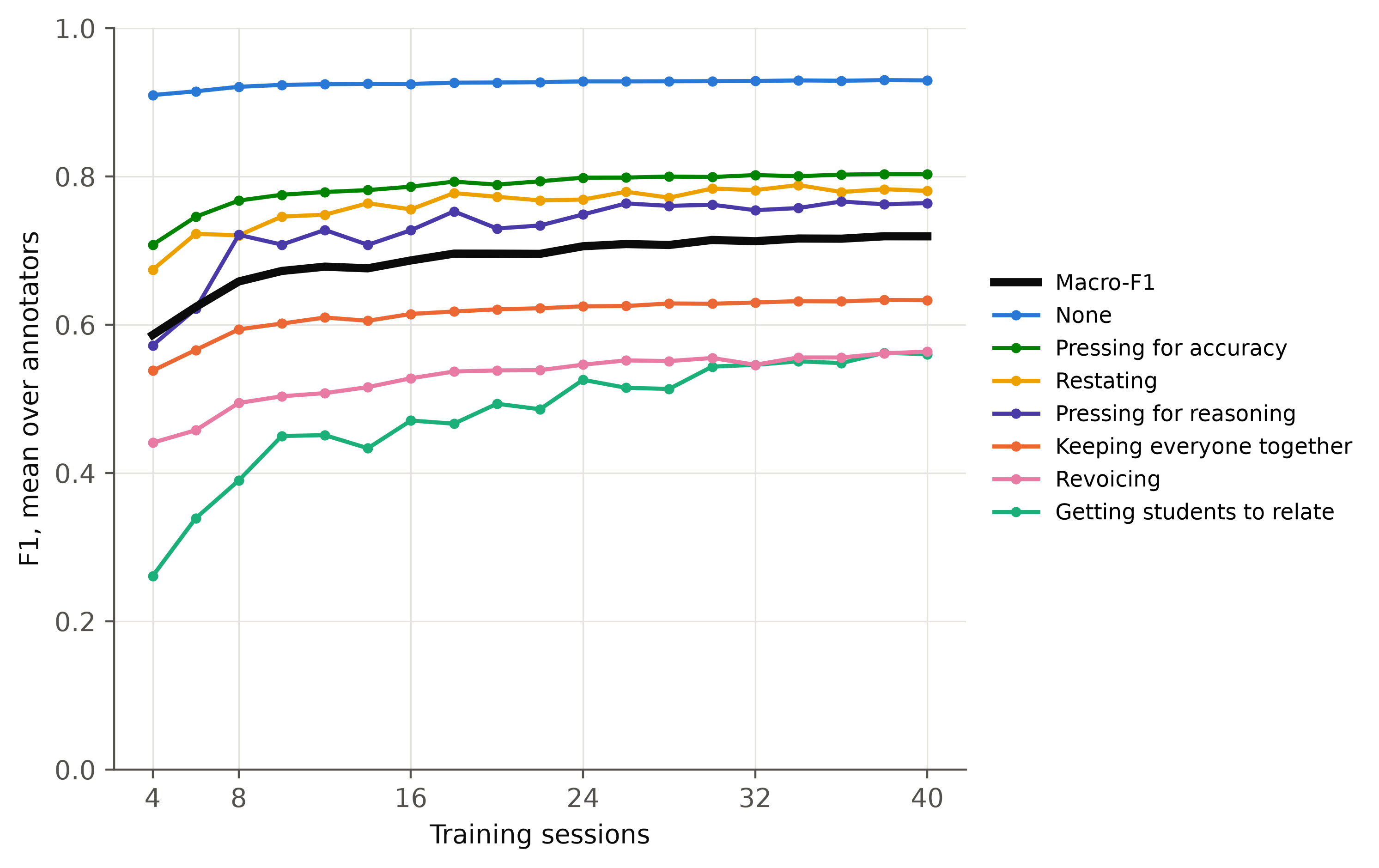}
    \caption{Per-move $F_1$ (coloured) and macro-$F_1$ (black) as the classifier is refitted on fewer training sessions, from 4 up to 38 of the 40 in each outer fold. Each test fold keeps its sessions, so every point is scored on the same utterances; each size is drawn twice, and points are the mean over the five annotators.}
    \label{fig:learning-curve-sessions}
\end{figure}

\begin{figure}[t]
    \centering
    \includegraphics[width=\columnwidth]{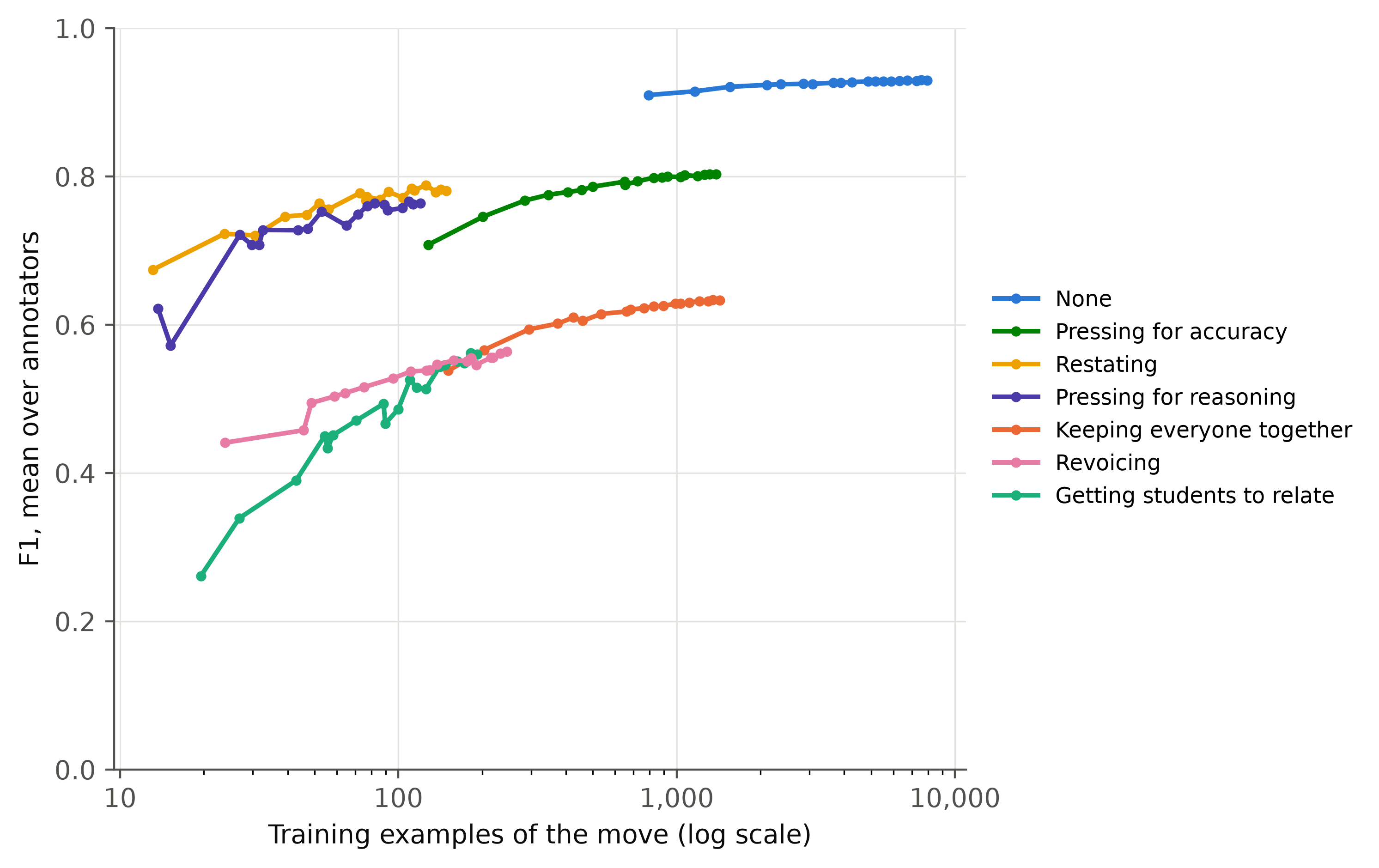}
    \caption{The same refits as Figure~\ref{fig:learning-curve-sessions}, with each move's $F_1$ plotted against the number of training utterances of that move (log scale), averaged over test folds. Points are the mean over the five annotators.}
    \label{fig:learning-curve-positives}
\end{figure}

 We investigate question: \textit{``How much data do I need to make this work?"} In Figure~\ref{fig:learning-curve-sessions} we show how performance changes as the top-rung classifier is trained on fewer sessions. Each test fold keeps its sessions, so every point is scored on the same utterances. We see that some labels' individual $F_1$ scores are relatively flat and achieve an $F_1$ close their final score within 8 sessions, while other labels performance grows more steadily as the number of sessions rises to 40. However, the session-level viewpoint blunts the picture since some moves are much rarer than others. So in Figure~\ref{fig:learning-curve-positives} we show the same refits with each move's $F_1$ against the number of training examples of that move, showing that even if two labels have appeared the same number of times in the data, there is label-level variance in the number of training examples needed to make accurate predictions.

\par
\begin{figure}[ht]
    \centering
    \includegraphics[width=0.8\columnwidth]{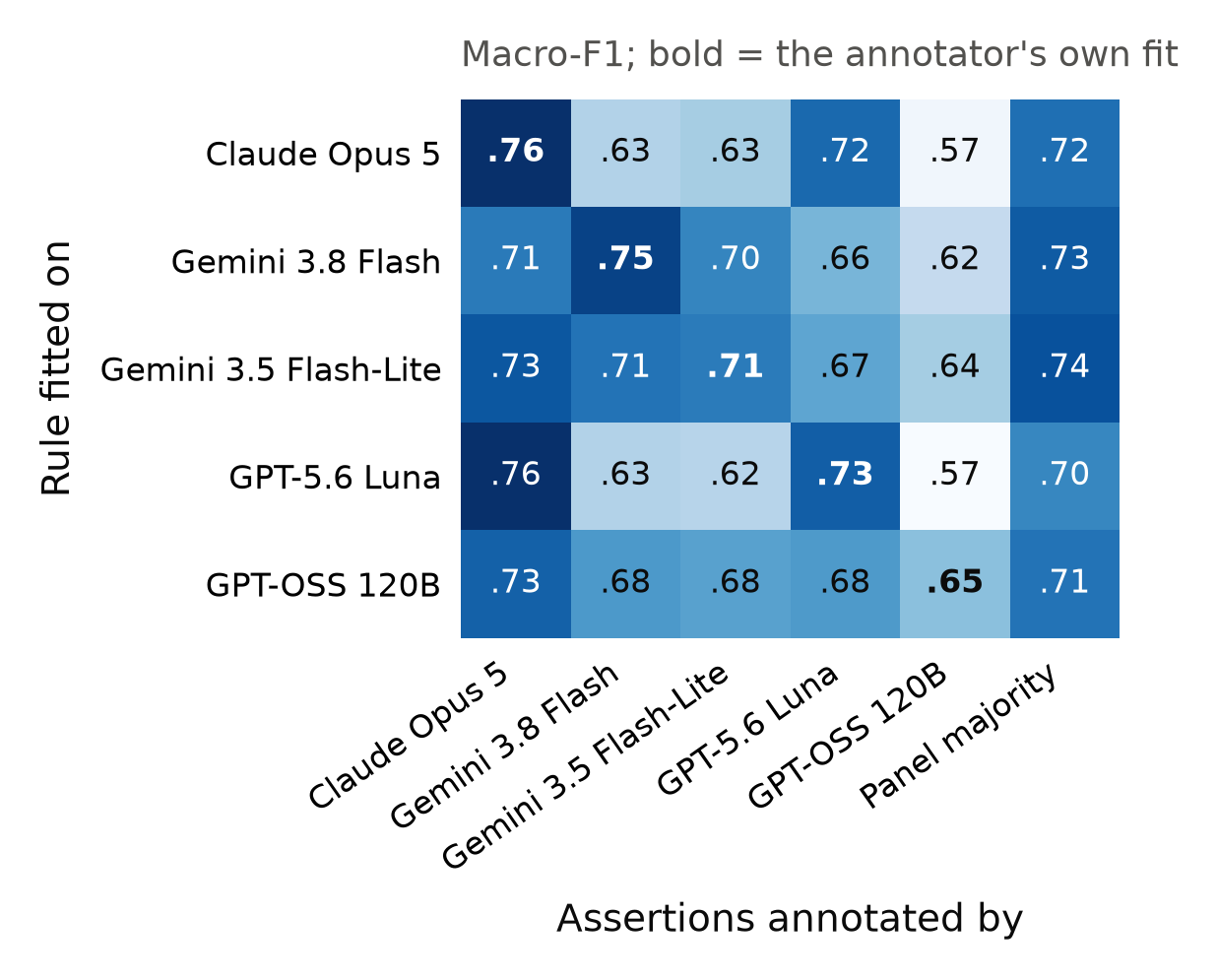}
    \caption{Macro-$F_1$ when a classifier fitted on one annotator's assertions (rows) is applied, unchanged, to another annotator's assertion values or to the panel majority's. Bold cells, on the diagonal, are each annotator's own cross-validated fit.}
    \label{fig:transfer}
\end{figure}

 In Figure~\ref{fig:transfer} we show what happens when a classifier fitted on one annotator's assertions is applied, unchanged, to another annotator's assertion values or to the panel majority's. The diagonal is each annotator's own fit. Overall, there is significant variability in the resulting Macro-F1s, although transfer remains high for our two most-agreeing annotators (Opus 5 and 5.6 Luna). 

\begin{table*}[!tp]
\centering
\small
\setlength{\tabcolsep}{3pt}
\begin{tabular}{lclrrrrrrrr}
\toprule
 &  &  &  & \multicolumn{7}{c}{F1 by talk move} \\
\cmidrule(lr){5-11}
Annotator & TD & Macro-F1 [95\% CI] & $\kappa$ & None & KET & GSR & PFA & PFR & RES & REV \\
\midrule
\textit{Words only} &  & 0.390 [0.365, 0.412] & 0.332 & 0.806 & 0.267 & 0.231 & 0.614 & 0.603 & 0.133 & 0.079 \\
\midrule
Claude Opus 5 &  & 0.695 [0.670, 0.716] & 0.700 & 0.936 & 0.596 & 0.553 & 0.771 & 0.774 & 0.686 & 0.549 \\
 & $\checkmark$ & 0.764 [0.738, 0.786] & \textbf{0.756} & \textbf{0.941} & \textbf{0.683} & \textbf{0.662} & 0.837 & 0.748 & \textbf{0.843} & 0.631 \\
\addlinespace[3pt]
Gemini 3.8 Flash &  & 0.673 [0.648, 0.695] & 0.686 & 0.933 & 0.586 & 0.515 & 0.750 & 0.731 & 0.667 & 0.531 \\
 & $\checkmark$ & 0.750 [0.724, 0.771] & 0.731 & 0.938 & 0.644 & 0.628 & 0.799 & 0.780 & 0.820 & \textbf{0.638} \\
\addlinespace[3pt]
Gemini 3.5 Flash-Lite &  & 0.640 [0.613, 0.661] & 0.644 & 0.918 & 0.557 & 0.435 & 0.737 & 0.785 & 0.607 & 0.440 \\
 & $\checkmark$ & 0.706 [0.676, 0.734] & 0.696 & 0.926 & 0.629 & 0.536 & 0.783 & 0.766 & 0.792 & 0.513 \\
\addlinespace[3pt]
GPT-5.6 Luna &  & 0.667 [0.642, 0.688] & 0.660 & 0.924 & 0.537 & 0.475 & 0.761 & 0.789 & 0.652 & 0.532 \\
 & $\checkmark$ & 0.728 [0.701, 0.750] & 0.722 & 0.932 & 0.642 & 0.561 & 0.822 & 0.769 & 0.790 & 0.581 \\
\addlinespace[3pt]
GPT-OSS 120B &  & 0.605 [0.578, 0.628] & 0.607 & 0.910 & 0.502 & 0.393 & 0.717 & 0.723 & 0.602 & 0.389 \\
 & $\checkmark$ & 0.649 [0.620, 0.673] & 0.648 & 0.913 & 0.569 & 0.415 & 0.774 & 0.759 & 0.657 & 0.456 \\
\midrule
\multicolumn{11}{l}{\textit{Combined annotators}} \\
\addlinespace[2pt]
All &  & 0.607 [0.578, 0.632] & 0.590 & 0.902 & 0.497 & 0.379 & 0.707 & 0.738 & 0.648 & 0.378 \\
 & $\checkmark$ & 0.649 [0.620, 0.674] & 0.633 & 0.908 & 0.550 & 0.424 & 0.753 & 0.754 & 0.720 & 0.434 \\
\addlinespace[3pt]
Any &  & 0.694 [0.670, 0.715] & 0.708 & 0.936 & 0.616 & 0.568 & 0.780 & 0.776 & 0.652 & 0.532 \\
 & $\checkmark$ & \textbf{0.765} [0.741, 0.784] & 0.753 & 0.939 & 0.679 & 0.651 & \textbf{0.838} & \textbf{0.794} & 0.821 & 0.630 \\
\addlinespace[3pt]
Majority &  & 0.668 [0.643, 0.688] & 0.671 & 0.929 & 0.555 & 0.512 & 0.754 & 0.759 & 0.650 & 0.514 \\
 & $\checkmark$ & 0.736 [0.712, 0.757] & 0.721 & 0.932 & 0.639 & 0.590 & 0.812 & 0.769 & 0.827 & 0.585 \\
\bottomrule
\end{tabular}
\caption{TalkMove classification across the fit grid. Each row is one configuration: lexical features plus every corpus-derived assertion, and, where marked (TD), every construct-derived assertion. Words only uses neither. Macro-F1 is shown with its 95\% session-bootstrap interval; $\kappa$ is Cohen's kappa. The best value in each column is in bold.}
\label{tab:fit-grid}
\par\smallskip
\end{table*}

\FloatBarrier

\section{Interpretability}
\label{sec:appendix interpretability}
One benefit of the EduBehaviors framework is that our fitted classifiers have an inspectable linear form. As such, we can inspect fitted coefficients corresponding to different classification heads to understand what evidence our classifier is using to distinguish between classes. However, our classification schemes regularize heads at different strengths. As such, we re-fit the classifiers from the main analysis at a shared penalty to investigate coefficient differences. We pair four of the TalkMoves into labels with potential overlap (PFA, PFR, and RES/REV). Then, for each of these pairs, we take the 15 assertions whose coefficients have the greatest absolute difference between the two classification heads.
\par
\begin{figure*}[p]
    \centering
    \includegraphics[width=0.9\textwidth]{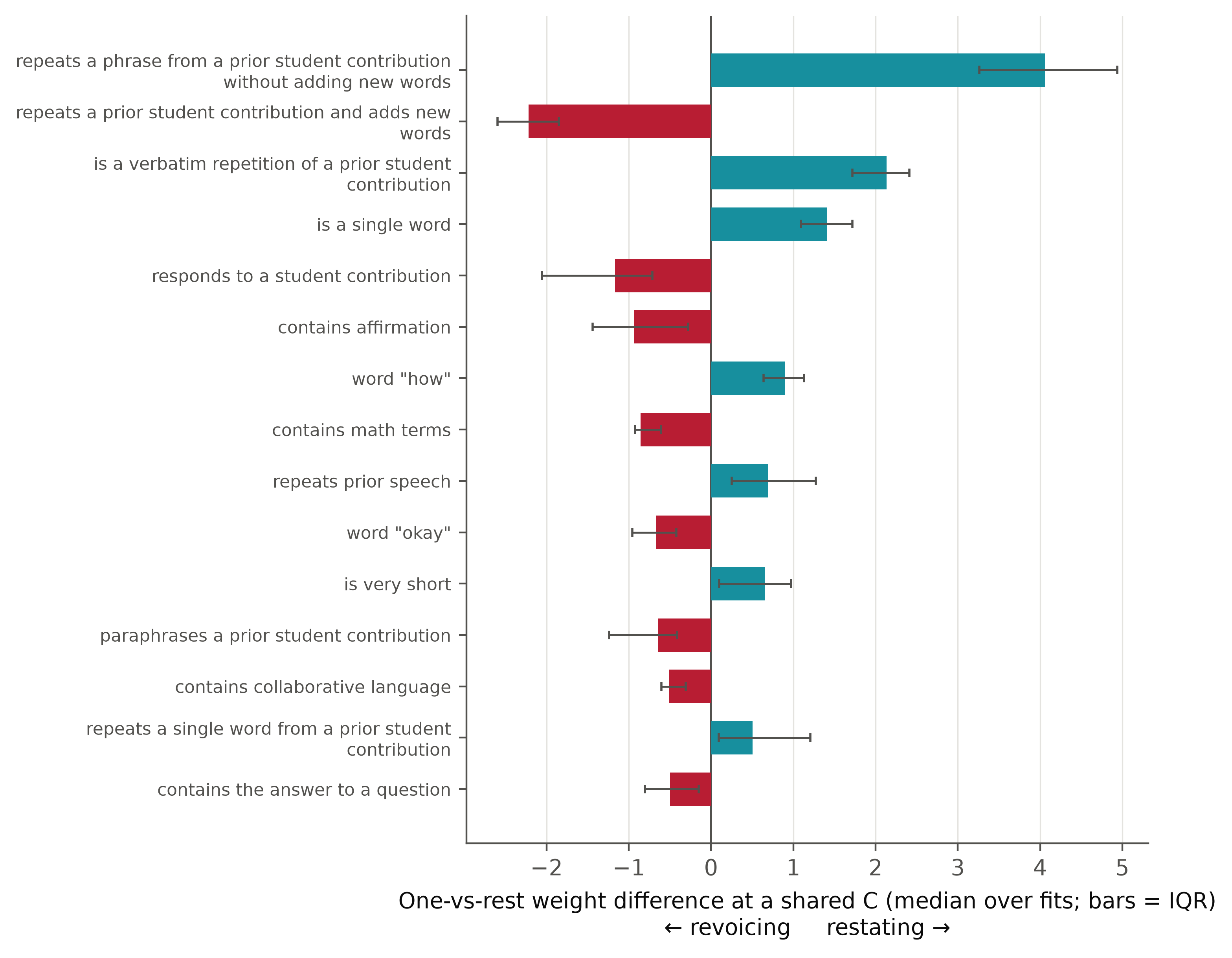}
    \caption{What separates Restating from Revoicing in the main classifier: the two moves' one-vs-rest weights at a shared $C$, differenced (median over folds and annotators; bars are the interquartile range).}
    \label{fig:contrast-res-rev}
\end{figure*}

\begin{figure*}[p]
    \centering
    \includegraphics[width=0.9\textwidth]{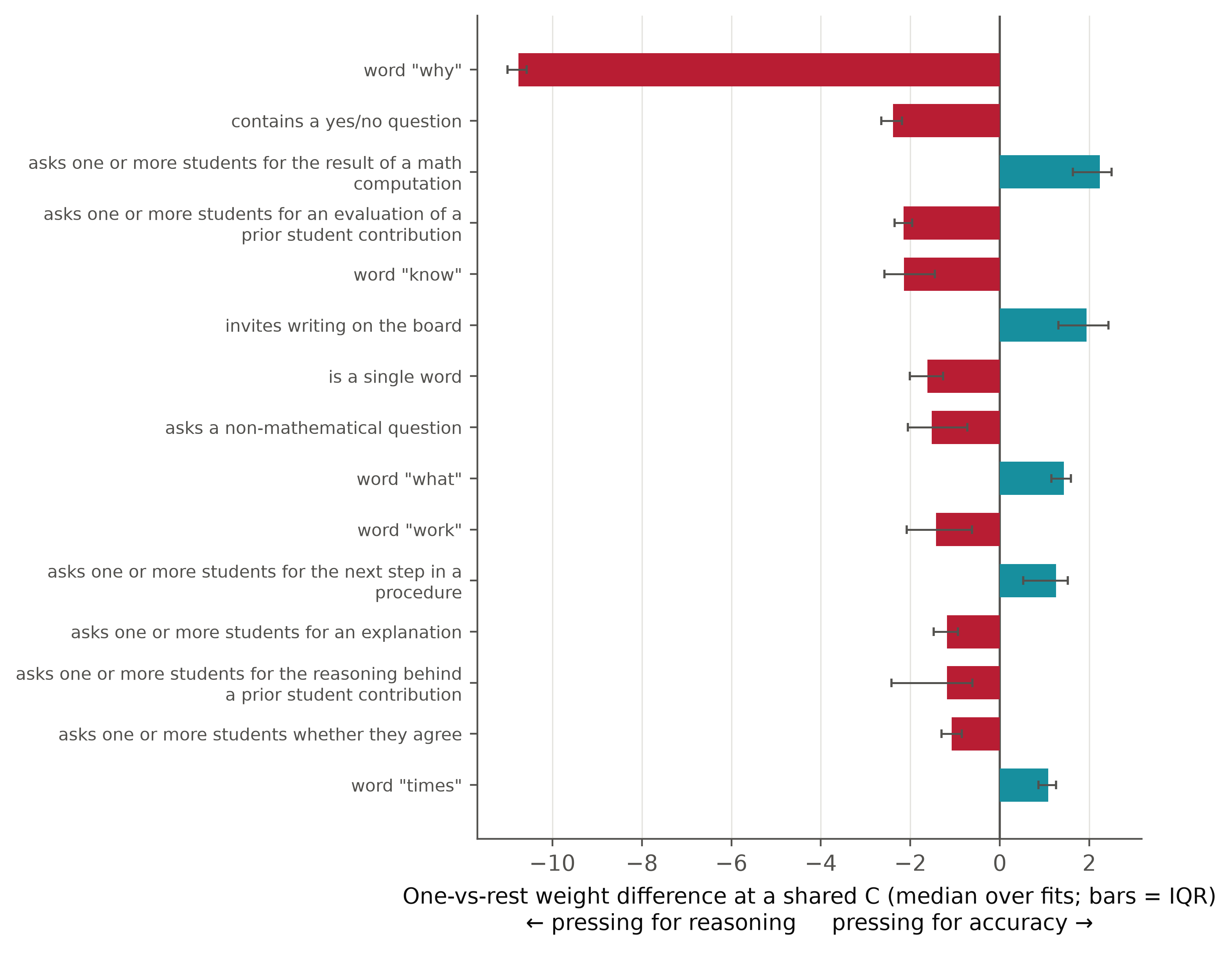}
    \caption{As Figure~\ref{fig:contrast-res-rev}, for Pressing for accuracy and Pressing for reasoning.}
    \label{fig:contrast-pfa-pfr}
\end{figure*}

 In Figure~\ref{fig:contrast-res-rev} we show what separates Restating from Revoicing in the main classifier: for each feature, the difference between its weights for the two moves. Features extending to the left distinguish revoicing from restating, and those on the right distinguish restating from revoicing. We see that the top restating-distinguishing feature is ``The utterance repeats a phrase from a prior student contribution without adding new words," showing that the logistic regression identifies a definitional aspect of the label. In Figure~\ref{fig:contrast-pfa-pfr} we show the same contrast for Pressing for accuracy and Pressing for reasoning. We see that even our word-based assertions play a distinguishing role here: the assertion most distinguishing PFR from PFA is simply the presence of the word ``why," matching the codebook's definition.

\FloatBarrier

\section{Error Analysis}
\label{sec:appendix errors}

\par
\begin{figure}[ht]
    \centering
    \includegraphics[width=0.8\columnwidth]{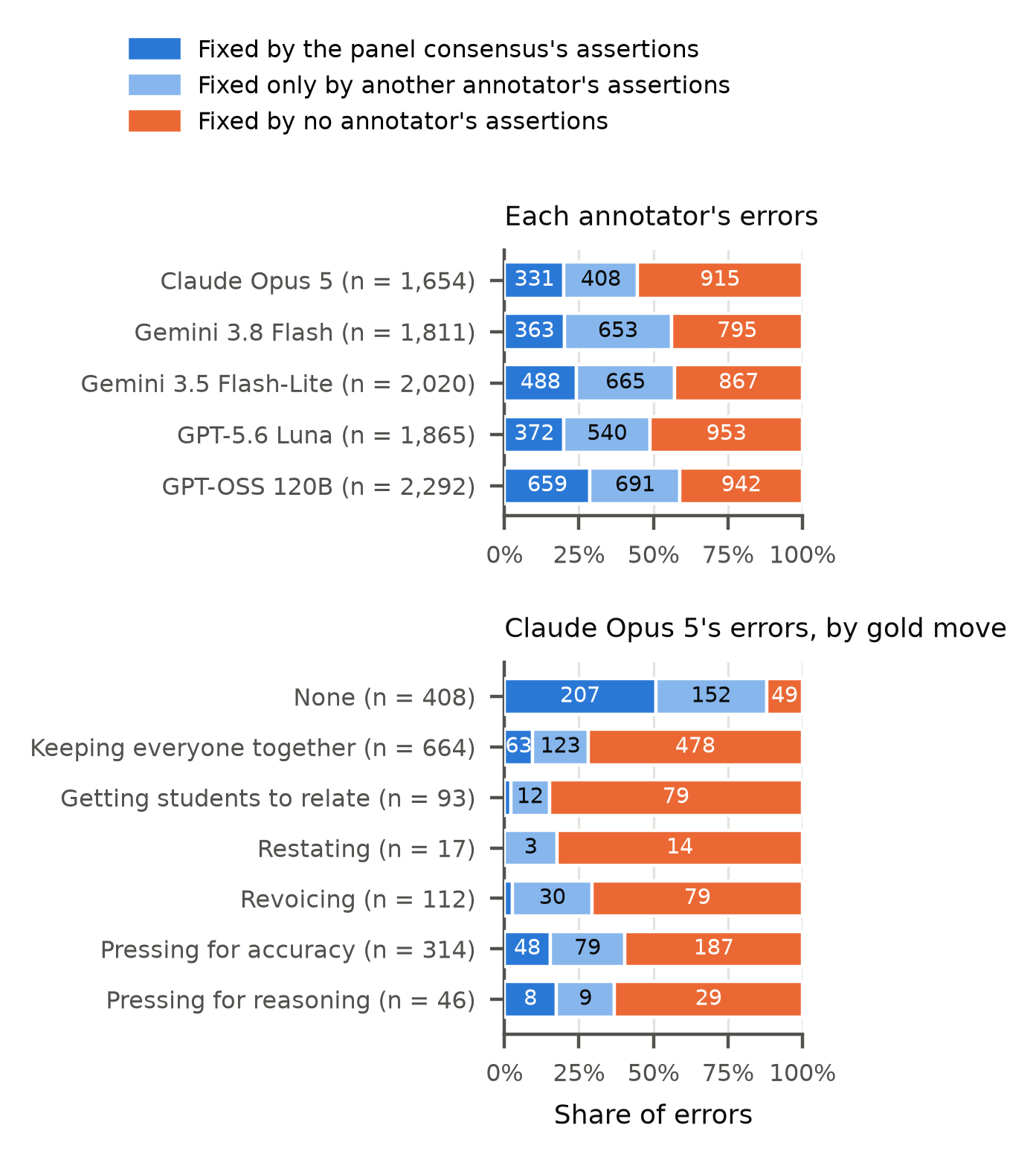}
    \caption{Each classification error, by what fixes it. The same held-out classifier is rescored with other assertion values: the panel consensus's (at least three of five annotators), then each other annotator's. An error is fixed by the consensus if those values give the gold move; fixed only by another annotator if the consensus's do not but some annotator's do; and fixed by no annotator otherwise. Top: every annotator. Bottom: Claude Opus 5, by gold move.}
    \label{fig:error-outcomes}
\end{figure}

 As discussed in the main paper, the framework allows errors to be characterized as annotation noise versus insufficiency of the current set of assertions and the rule combining them. To that end, we define three error categories: fixable by consensus, fixable by annotation swap, and those that remain. Errors fixable by consensus would be resolved if that utterance took the majority annotations for some assertion, errors fixable by annotation swap would be resolved if that utterance took on the annotations from some other specific annotator. Both of these point to iterating on assertion annotation, but the former implies that the panel of annotators can identify this and the error may have been idiosyncratic to this annotator and can be resolved by a minor rewrite of the assertion. The second type of error implies that there is \textit{some} model whose observations would fix the error but this understanding is model-contingent. As such, this might imply a more effortful rewriting or re-definition of the assertion. For errors not falling into either bucket, increased performance can only come from adding new assertions or choosing a new rule function. In Figure~\ref{fig:error-outcomes} we show the distribution of the three error-types among the errors contributing to the main results and zoom into the per-label distribution. We see that for each annotator the third bucket contains the most errors, but after looking at the per-label distributions for Claude Opus 5, we see that the ``None" class has a significant degree of errors in the first two buckets.
\par By swapping our focus from the annotator-level analysis to an assertion-level analysis, we can see which assertions contribute to the most errors that are fixable by consensus. In Table~\ref{tab:decisive-assertions} we show the 10 assertions contribution to the greatest number of fixable by consensus errors. This view goes beyond the task-independent view given by checking cross-model agreement and highlights the assertion edits that will be most consequential for downstream classification.


\begin{table*}[t]
\centering
\small
\begin{tabular}{lrrr}
\toprule
Assertion & $\alpha$ & Fire rate & Errors \\
\midrule
invites participation (BU) & 0.66 & 27.4\% & 327 \\
contains a yes/no question (TD) & 0.80 & 8.3\% & 288 \\
contains a question (BU) & 0.85 & 21.4\% & 222 \\
invites a student to speak next (TD) & 0.53 & 16.0\% & 208 \\
repeats a student answer (BU) & 0.64 & 3.6\% & 155 \\
invites student thinking (BU) & 0.61 & 15.4\% & 136 \\
responds to a student contribution (BU) & 0.53 & 18.9\% & 130 \\
repeats a prior student contribution and adds new words (TD) & 0.39 & 2.0\% & 84 \\
asks one or more students for an evaluation of a prior student contribution (TD) & 0.56 & 1.0\% & 71 \\
is a declarative statement (BU) & 0.64 & 39.6\% & 68 \\
\bottomrule
\end{tabular}
\caption{The 10 assertions that decide the most errors: errors, over all five annotators, that flipping the assertion alone from the annotator's value to the panel consensus's value fixes under the same head. ``The utterance'' is omitted; BU = corpus-derived, TD = construct-derived.}
\label{tab:decisive-assertions}
\end{table*}


\begin{thebibliography}{34}
\providecommand{\natexlab}[1]{#1}

\bibitem[{Agrawal et~al.(2025)Agrawal, Ahemad, and
  Sembium}]{agrawal2025rationale}
Sanjay Agrawal, Faizan Ahemad, and Vivek~Varadarajan Sembium. 2025.
\newblock Rationale-guided distillation for e-commerce relevance
  classification: Bridging large language models and lightweight
  cross-encoders.
\newblock In \emph{Proceedings of the 31st International Conference on
  Computational Linguistics: Industry Track}, pages 136--148.

\bibitem[{Ahtisham et~al.(2026)Ahtisham, Vanacore, Zhou, Lee, and
  Kizilcec}]{ahtisham2026llmreasoningpredictsmodels}
Bakhtawar Ahtisham, Kirk Vanacore, Zhuqian Zhou, Jinsook Lee, and Rene~F.
  Kizilcec. 2026.
\newblock \href {https://arxiv.org/abs/2602.09832} {Llm reasoning predicts when
  models are right: Evidence from coding classroom discourse}.
\newblock \emph{Preprint}, arXiv:2602.09832.

\bibitem[{Borchers et~al.(2025)Borchers, Thomas, Lin, Abboud, and
  Koedinger}]{borchers2025distillation}
Conrad Borchers, Danielle~R. Thomas, Jionghao Lin, Ralph Abboud, and Kenneth~R.
  Koedinger. 2025.
\newblock \href {https://ceur-ws.org/Vol-3994/short1.pdf} {Augmenting
  human-annotated training data with large language model generation and
  distillation in open-response assessment}.
\newblock In \emph{Proceedings of the Second International Workshop on
  Generative AI for Learning Analytics}, volume 3994 of \emph{CEUR Workshop
  Proceedings}, pages 42--50. CEUR-WS.org.

\bibitem[{Dell and Rambachan(2026)}]{dell2026measurement}
Melissa Dell and Ashesh Rambachan. 2026.
\newblock \href {https://doi.org/10.3386/w35744} {The measurement revolution?
  credible measurement and inference in the age of {AI}}.
\newblock Working Paper 35744, National Bureau of Economic Research.

\bibitem[{Demszky and Hill(2023)}]{demszky-hill-2023-ncte}
Dorottya Demszky and Heather Hill. 2023.
\newblock \href {https://doi.org/10.18653/v1/2023.bea-1.44} {The {NCTE}
  transcripts: A dataset of elementary math classroom transcripts}.
\newblock In \emph{Proceedings of the 18th Workshop on Innovative Use of NLP
  for Building Educational Applications (BEA 2023)}, pages 528--538, Toronto,
  Canada. Association for Computational Linguistics.

\bibitem[{Demszky et~al.(2021)Demszky, Liu, Mancenido, Cohen, Hill, Jurafsky,
  and Hashimoto}]{demszky-etal-2021-measuring}
Dorottya Demszky, Jing Liu, Zid Mancenido, Julie Cohen, Heather Hill, Dan
  Jurafsky, and Tatsunori Hashimoto. 2021.
\newblock \href {https://doi.org/10.18653/v1/2021.acl-long.130} {Measuring
  conversational uptake: A case study on student-teacher interactions}.
\newblock In \emph{Proceedings of the 59th Annual Meeting of the Association
  for Computational Linguistics and the 11th International Joint Conference on
  Natural Language Processing (Volume 1: Long Papers)}, pages 1638--1653,
  Online. Association for Computational Linguistics.

\bibitem[{Gilardi et~al.(2023)Gilardi, Alizadeh, and
  Kubli}]{gilardi-etal-2023-chatgpt}
Fabrizio Gilardi, Meysam Alizadeh, and Ma{\"e}l Kubli. 2023.
\newblock \href {https://doi.org/10.1073/pnas.2305016120} {{ChatGPT}
  outperforms crowd workers for text-annotation tasks}.
\newblock \emph{Proceedings of the National Academy of Sciences},
  120(30):e2305016120.

\bibitem[{Halterman and Keith(2025)}]{halterman-keith-2025-codebook}
Andrew Halterman and Katherine~A. Keith. 2025.
\newblock \href {https://doi.org/10.1017/pan.2025.10017} {Codebook {LLMs}:
  Evaluating {LLMs} as measurement tools for political science concepts}.
\newblock \emph{Political Analysis}, pages 1--17.

\bibitem[{He et~al.(2025)He, Naphade, and Huang}]{he-etal-2025-prompting}
Zeyu He, Saniya Naphade, and Ting-Hao~Kenneth Huang. 2025.
\newblock \href {https://doi.org/10.1145/3706598.3714319} {Prompting in the
  dark: Assessing human performance in prompt engineering for data labeling
  when gold labels are absent}.
\newblock In \emph{Proceedings of the 2025 CHI Conference on Human Factors in
  Computing Systems}. Association for Computing Machinery.

\bibitem[{Hedley et~al.(2026)Hedley, Pietrzak, Dias, Burden, Ahtisham, Zhou,
  Vanacore, Marland, Slama, Reich, Koedinger, and
  Kizilcec}]{hedley-etal-2026-sandpiper}
Daryl Hedley, Doug Pietrzak, Jorge Dias, Ian Burden, Bakhtawar Ahtisham,
  Zhuqian Zhou, Kirk Vanacore, Josh Marland, Rachel Slama, Justin Reich,
  Kenneth Koedinger, and Ren{\'e} Kizilcec. 2026.
\newblock \href {https://doi.org/10.48550/arXiv.2603.08406} {Sandpiper:
  Orchestrated {AI}-annotation for educational discourse at scale}.
\newblock \emph{Preprint}, arXiv:2603.08406.

\bibitem[{Khot et~al.(2023)Khot, Trivedi, Finlayson, Fu, Richardson, Clark, and
  Sabharwal}]{khot-etal-2023-decomposed}
Tushar Khot, Harsh Trivedi, Matthew Finlayson, Yao Fu, Kyle Richardson, Peter
  Clark, and Ashish Sabharwal. 2023.
\newblock \href {https://openreview.net/forum?id=_nGgzQjzaRy} {Decomposed
  prompting: A modular approach for solving complex tasks}.
\newblock In \emph{Proceedings of the 11th International Conference on Learning
  Representations}.

\bibitem[{Koh et~al.(2020)Koh, Nguyen, Tang, Mussmann, Pierson, Kim, and
  Liang}]{koh2020concept}
Pang~Wei Koh, Thao Nguyen, Yew~Siang Tang, Stephen Mussmann, Emma Pierson, Been
  Kim, and Percy Liang. 2020.
\newblock Concept bottleneck models.
\newblock In \emph{International conference on machine learning}, pages
  5338--5348. Pmlr.

\bibitem[{Lam et~al.(2024)Lam, Teoh, Landay, Heer, and
  Bernstein}]{lam2024lloom}
Michelle~S. Lam, Janice Teoh, James Landay, Jeffrey Heer, and Michael~S.
  Bernstein. 2024.
\newblock \href {https://doi.org/10.1145/3613904.3642830} {Concept induction:
  Analyzing unstructured text with high-level concepts using {LLooM}}.
\newblock In \emph{Proceedings of the CHI Conference on Human Factors in
  Computing Systems}. Association for Computing Machinery.

\bibitem[{Liu et~al.(2025)Liu, Zambrano, Baker, Barany, Ocumpaugh, Zhang,
  Pankiewicz, Nasiar, and Wei}]{liu2025qualitative}
Xiner Liu, Andres~Felipe Zambrano, Ryan~S Baker, Amanda Barany, Jaclyn
  Ocumpaugh, Jiayi Zhang, Maciej Pankiewicz, Nidhi Nasiar, and Zhanlan Wei.
  2025.
\newblock Qualitative coding with gpt-4: Where it works better.
\newblock \emph{Journal of Learning Analytics}, 12(1).

\bibitem[{Long et~al.(2024)Long, Luo, and Zhang}]{long-etal-2024-evaluating}
Yun Long, Haifeng Luo, and Yu~Zhang. 2024.
\newblock \href {https://doi.org/10.1038/s41539-024-00273-3} {Evaluating large
  language models in analysing classroom dialogue}.
\newblock \emph{npj Science of Learning}, 9(60).

\bibitem[{Ludan et~al.(2023)Ludan, Lyu, Yang, Dugan, Yatskar, and
  Callison-Burch}]{ludan2023textbottleneck}
Josh~Magnus Ludan, Qing Lyu, Yue Yang, Liam Dugan, Mark Yatskar, and Chris
  Callison-Burch. 2023.
\newblock \href {https://doi.org/10.48550/arXiv.2310.19660}
  {Interpretable-by-design text understanding with iteratively generated
  concept bottleneck}.
\newblock \emph{Preprint}, arXiv:2310.19660.

\bibitem[{Michaels et~al.(2008)Michaels, O'Connor, and
  Resnick}]{michaels-etal-2008-deliberative}
Sarah Michaels, Catherine O'Connor, and Lauren~B. Resnick. 2008.
\newblock \href {https://doi.org/10.1007/s11217-007-9071-1} {Deliberative
  discourse idealized and realized: Accountable talk in the classroom and in
  civic life}.
\newblock \emph{Studies in Philosophy and Education}, 27(4):283--297.

\bibitem[{Moreau-Pernet et~al.(2024)Moreau-Pernet, Tian, Sawaya, Foltz, Cao,
  Milne, and Christie}]{moreau-pernet-etal-2024-classifying}
Baptiste Moreau-Pernet, Yu~Tian, Sandra Sawaya, Peter Foltz, Jie Cao, Brent
  Milne, and Thomas Christie. 2024.
\newblock \href {https://doi.org/10.1145/3657604.3664664} {Classifying tutor
  discursive moves at scale in mathematics classrooms with large language
  models}.
\newblock In \emph{Proceedings of the Eleventh ACM Conference on Learning @
  Scale}, pages 361--365, Atlanta, GA, USA. Association for Computing
  Machinery.

\bibitem[{Pangakis et~al.(2023)Pangakis, Wolken, and
  Fasching}]{pangakis-etal-2023-validation}
Nicholas Pangakis, Samuel Wolken, and Neil Fasching. 2023.
\newblock \href {https://doi.org/10.48550/arXiv.2306.00176} {Automated
  annotation with generative {AI} requires validation}.
\newblock \emph{Preprint}, arXiv:2306.00176.

\bibitem[{Pryzant et~al.(2023)Pryzant, Iter, Li, Lee, Zhu, and
  Zeng}]{pryzant-etal-2023-automatic}
Reid Pryzant, Dan Iter, Jerry Li, Yin Lee, Chenguang Zhu, and Michael Zeng.
  2023.
\newblock \href {https://doi.org/10.18653/v1/2023.emnlp-main.494} {Automatic
  prompt optimization with ``gradient descent'' and beam search}.
\newblock In \emph{Proceedings of the 2023 Conference on Empirical Methods in
  Natural Language Processing}, pages 7957--7968, Singapore. Association for
  Computational Linguistics.

\bibitem[{Ratner et~al.(2020)Ratner, Bach, Ehrenberg, Fries, Wu, and
  R{\'e}}]{ratner-etal-2020-snorkel}
Alexander Ratner, Stephen~H. Bach, Henry Ehrenberg, Jason Fries, Sen Wu, and
  Christopher R{\'e}. 2020.
\newblock \href {https://doi.org/10.1007/s00778-019-00552-1} {Snorkel: Rapid
  training data creation with weak supervision}.
\newblock \emph{The VLDB Journal}, 29(2):709--730.

\bibitem[{Reimers and Gurevych(2019)}]{reimers-2019-sentence-bert}
Nils Reimers and Iryna Gurevych. 2019.
\newblock \href {http://arxiv.org/abs/1908.10084} {Sentence-bert: Sentence
  embeddings using siamese bert-networks}.
\newblock In \emph{Proceedings of the 2019 Conference on Empirical Methods in
  Natural Language Processing}. Association for Computational Linguistics.

\bibitem[{Reiss(2023)}]{reiss-2023-reliability}
Michael~V. Reiss. 2023.
\newblock \href {https://doi.org/10.48550/arXiv.2304.11085} {Testing the
  reliability of {ChatGPT} for text annotation and classification: A cautionary
  remark}.
\newblock \emph{Preprint}, arXiv:2304.11085.

\bibitem[{Shankar et~al.(2024)Shankar, Zamfirescu-Pereira, Hartmann,
  Parameswaran, and Arawjo}]{shankar2024validators}
Shreya Shankar, J.~D. Zamfirescu-Pereira, Bj{\"o}rn Hartmann, Aditya~G.
  Parameswaran, and Ian Arawjo. 2024.
\newblock \href {https://doi.org/10.1145/3654777.3676450} {Who validates the
  validators? aligning {LLM}-assisted evaluation of {LLM} outputs with human
  preferences}.
\newblock In \emph{Proceedings of the 37th Annual ACM Symposium on User
  Interface Software and Technology}. Association for Computing Machinery.

\bibitem[{Suresh et~al.(2022)Suresh, Jacobs, Harty, Perkoff, Martin, and
  Sumner}]{suresh-etal-2022-talkmoves}
Abhijit Suresh, Jennifer Jacobs, Charis Harty, Margaret Perkoff, James~H.
  Martin, and Tamara Sumner. 2022.
\newblock \href {https://aclanthology.org/2022.lrec-1.497/} {The {TalkMoves}
  dataset: K-12 mathematics lesson transcripts annotated for teacher and
  student discursive moves}.
\newblock In \emph{Proceedings of the Thirteenth Language Resources and
  Evaluation Conference}, pages 4654--4662, Marseille, France. European
  Language Resources Association.

\bibitem[{Thomas et~al.(2025)Thomas, Borchers, Lin, Kakarla, Bhushan, Gatz,
  Gupta, Abboud, and Koedinger}]{thomas2025leveraging}
Danielle~R Thomas, Conrad Borchers, Jionghao Lin, Sanjit Kakarla, Shambhavi
  Bhushan, Erin Gatz, Shivang Gupta, Ralph Abboud, and Kenneth~R Koedinger.
  2025.
\newblock Leveraging llms to assess tutor moves in real-life dialogues: A
  feasibility study.
\newblock In \emph{European Conference on Technology Enhanced Learning}, pages
  268--273. Springer.

\bibitem[{Tunstall et~al.(2022)Tunstall, Reimers, Jo, Bates, Korat, Wasserblat,
  and Pereg}]{setfit}
Lewis Tunstall, Nils Reimers, Unso Eun~Seo Jo, Luke Bates, Daniel Korat, Moshe
  Wasserblat, and Oren Pereg. 2022.
\newblock \href {https://doi.org/10.48550/ARXIV.2209.11055} {Efficient few-shot
  learning without prompts}.
\newblock \emph{arXiv preprint}.

\bibitem[{Vanacore and Kizilcec(2025)}]{vanacore-kizilcec-2025-instructional}
Kirk Vanacore and Rene~F. Kizilcec. 2025.
\newblock \href {https://doi.org/10.48550/arXiv.2512.19903} {How well do large
  language models recognize instructional moves? establishing baselines for
  foundation models in educational discourse}.
\newblock \emph{Preprint}, arXiv:2512.19903.

\bibitem[{Wang and Demszky(2024)}]{wang2024educonvokit}
Rose~E. Wang and Dorottya Demszky. 2024.
\newblock \href {https://doi.org/10.18653/v1/2024.naacl-demo.6}
  {{Edu-ConvoKit}: An open-source library for education conversation data}.
\newblock In \emph{Proceedings of the 2024 Conference of the North American
  Chapter of the Association for Computational Linguistics: Human Language
  Technologies (Volume 3: System Demonstrations)}, pages 61--69. Association
  for Computational Linguistics.

\bibitem[{Wang et~al.(2025)Wang, Ribeiro, Robinson, Loeb, and
  Demszky}]{wang-etal-2025-tutor}
Rose~E. Wang, Ana~Trindade Ribeiro, Carly~D. Robinson, Susanna Loeb, and
  Dorottya Demszky. 2025.
\newblock \href {https://doi.org/10.26300/81nh-8262} {Tutor {CoPilot}: A
  human-{AI} approach for scaling real-time expertise}.
\newblock EdWorkingPaper 24-1056, Annenberg Institute at Brown University.

\bibitem[{Xiong et~al.(2025)Xiong, Ni, Fan, Zouhar, Rooein,
  Calvo-Bartolom{\'e}, Hoyle, Jin, Sachan, Leippold, Hovy, El-Assady, and
  Ash}]{xiong-etal-2025-co}
Chenfei Xiong, Jingwei Ni, Yu~Fan, Vil{\'e}m Zouhar, Donya Rooein, Lorena
  Calvo-Bartolom{\'e}, Alexander Hoyle, Zhijing Jin, Mrinmaya Sachan, Markus
  Leippold, Dirk Hovy, Mennatallah El-Assady, and Elliott Ash. 2025.
\newblock \href {https://doi.org/10.18653/v1/2025.emnlp-demos.25} {Co-{DETECT}:
  Collaborative discovery of edge cases in text classification}.
\newblock In \emph{Proceedings of the 2025 Conference on Empirical Methods in
  Natural Language Processing: System Demonstrations}, pages 354--364, Suzhou,
  China. Association for Computational Linguistics.

\bibitem[{Ye et~al.(2024)Ye, Kim, Kim, Hwang, Kim, Jo, Thorne, Kim, and
  Seo}]{ye-etal-2024-flask}
Seonghyeon Ye, Doyoung Kim, Sungdong Kim, Hyeonbin Hwang, Seungone Kim, Yongrae
  Jo, James Thorne, Juho Kim, and Minjoon Seo. 2024.
\newblock \href {https://openreview.net/forum?id=CYmF38ysDa} {{FLASK}:
  Fine-grained language model evaluation based on alignment skill sets}.
\newblock In \emph{Proceedings of the 12th International Conference on Learning
  Representations}.

\bibitem[{Zambrano et~al.(2026)Zambrano, Wei, Zhang, Baker, Ocumpaugh, Barany,
  Liu, Zhou, Paquette, Ginger, and Borchers}]{zambrano2026codebook}
Andres~Felipe Zambrano, Zhanlan Wei, Jiayi Zhang, Ryan~S. Baker, Jaclyn
  Ocumpaugh, Amanda Barany, Xiner Liu, Yiqiu Zhou, Luc Paquette, Jeffrey
  Ginger, and Conrad Borchers. 2026.
\newblock \href {https://doi.org/10.5281/zenodo.18352290} {Data plus theory
  equals codebook: Leveraging {LLMs} for human-{AI} codebook development}.
\newblock \emph{Journal of Educational Data Mining}, 18(1):25--65.

\bibitem[{Ziems et~al.(2024)Ziems, Held, Shaikh, Chen, Zhang, and
  Yang}]{ziems-etal-2024-large}
Caleb Ziems, William Held, Omar Shaikh, Jiaao Chen, Zhehao Zhang, and Diyi
  Yang. 2024.
\newblock \href {https://doi.org/10.1162/coli_a_00502} {Can large language
  models transform computational social science?}
\newblock \emph{Computational Linguistics}, 50(1):237--291.

\end{thebibliography}
\end{document}